\documentclass[11pt]{article}

\usepackage[preprint]{acl}

\usepackage{times}
\usepackage{latexsym}
\usepackage[T1]{fontenc}
\usepackage[utf8]{inputenc}
\usepackage{microtype}
\usepackage{inconsolata}

\usepackage{graphicx}
\usepackage{booktabs}
\usepackage{amsmath}
\usepackage{amsfonts}
\usepackage{amssymb}
\usepackage{pifont}
\usepackage{bm}
\usepackage{array}
\usepackage{multirow}
\usepackage[table]{xcolor}
\usepackage{pgfplots}
\usepackage{cleveref}
\usepackage{url}
\usepackage{hyperref}
\usepackage{placeins}
\pgfplotsset{compat=1.18}

\usepackage{fontawesome5}
\definecolor{githubblue}{HTML}{0969DA}

\title{ViCoStream: Streaming VideoLLMs Can Run Beyond 100 FPS with Stage-Wise Coordinated Inference}

\author{
 \textbf{Yang Tan\textsuperscript{1}\footnotemark[2]},
 \textbf{Junlong Tong\textsuperscript{2,3}\footnotemark[2]},
 \textbf{Linan Yue\textsuperscript{1}},
 \textbf{Hao Wu\textsuperscript{2}},
 \textbf{Pengfei Fang\textsuperscript{1}\footnotemark[1]},
 \textbf{Xiaoyu Shen\textsuperscript{2}\footnotemark[1]}
\\
 \textsuperscript{1}Southeast University~~
 \textsuperscript{2}Eastern Institute of Technology, Ningbo~~
 \textsuperscript{3}Shanghai Jiao Tong University \\
 \texttt{yangtan@seu.edu.cn}~~~~\texttt{jl-tong@sjtu.edu.cn} \\
 \texttt{fangpengfei@seu.edu.cn}~~~~\texttt{xyshen@eitech.edu.cn}
}

\begin{document}
\maketitle
\begingroup
\renewcommand{\thefootnote}{\fnsymbol{footnote}}
\footnotetext[1]{Corresponding authors.}
\footnotetext[2]{Equal contribution.}
\endgroup

\begin{abstract}
Streaming VideoLLMs must continuously process incoming video while maintaining low query latency, making both video-ingestion throughput and query-time responsiveness critical for real-time deployment. Existing methods largely focus on accelerating individual modules, such as visual encoding, token pruning, or KV-cache compression, but provide limited insight into whether the resulting system can sustain real-time streaming performance. We formulate streaming VideoLLM inference as a coordinated pipeline spanning visual preprocessing, visual encoding, token dropping, and LLM prefilling/decoding. Building on this formulation, we propose \textbf{ViCoStream} (\textbf{Vi}deo \textbf{Co}ordinated \textbf{Stream}ing), a stage-wise coordinated streaming framework that combines chunk-wise execution, CUDA-stream overlap, visual token control, bounded visual attention, and query-side retrieval to bound per-chunk computation and memory costs. We further provide a systematic study of bottleneck migration, revealing how chunk size, token retention, attention locality, and retrieval scope shape the throughput--accuracy trade-off. Experiments with Qwen2.5-VL-3B/7B-Instruct across multiple streaming benchmarks show that ViCoStream achieves 134 FPS video throughput and $<50$ms TTFT on a single A100 GPU while maintaining accuracy close to full-history baselines. We release our code at \textcolor{githubblue}{\faGithub}\ \href{https://github.com/EIT-NLP/StreamingLLM/tree/main/ViCoStream}{\textbf{\texttt{EIT-NLP/ViCoStream}}}.

\end{abstract}

\section{Introduction}

Large VideoLLMs have achieved remarkable success across diverse video understanding tasks~\cite{chatterjee2025memory,fu2025video,zhang2024llava,ma2026survey,zhao2025urbanvideo,mu2023embodiedgpt,xu2024drivegpt4,xu2025vlm,ma2024dolphins}. In many real-world deployments, however, videos arrive continuously as streams, while user queries may occur before the full video has been observed~\cite{chen2024videollm,qian2024streaming,tong2026static,lin2026speak}. A practical streaming VideoLLM must therefore both keep pace with incoming video and answer queries with low latency, making \emph{video-ingestion throughput} and \emph{query-time time-to-first-token} (TTFT) the two primary system objectives.

\begin{figure}[t]
    \centering
    \includegraphics[width=\linewidth]{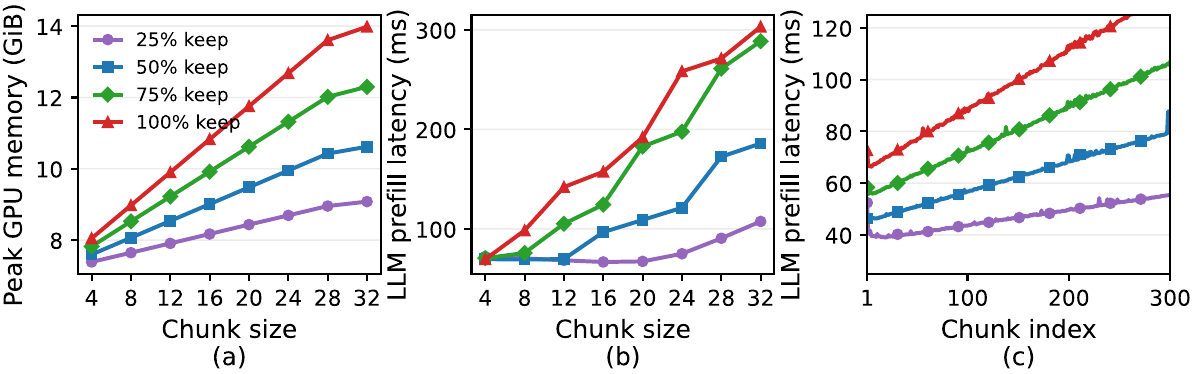}
    \vspace{-0.5cm}
    \caption{Memory and latency scaling under full-history streaming attention.
Token dropping reduces (a) peak GPU memory and (b) LLM prefill latency under simulated 720p streaming input, but (c) prefill latency still grows with the streaming chunk index when the visible visual history accumulates.
This suggests that token dropping alone cannot ensure long-duration latency stability without bounding the attended visual context.}
    \label{fig:streaming_pipelines}
    \vspace{-0.5cm}
\end{figure}

Existing streaming VideoLLMs generally adopt one of two paradigms (see Appendix~\ref{app:streaming_paradigms}). The first is \emph{delayed streaming}~\cite{dorovatas2026recurrent,yao2025timechat}, where incoming frames are accumulated but remain largely unprocessed until a user query arrives. At query time, the model performs a single large prefill over the entire observed history. While simple, this design causes query latency and peak GPU memory consumption to grow with stream length~\cite{wang2025stc,zhang2026hermes}. As a result, delayed streaming struggles to support real-time interaction over long videos.

The second paradigm is \emph{continuous streaming}, which incrementally processes video as frames arrive and continuously updates the model state. Recent work improves individual stages through sparse visual encoding~\cite{renggli2022learning,li2025videoscan,chen2024videollm}, token pruning and keyframe selection~\cite{kim2024token,wu2026data}, and memory-bank or KV-cache compression~\cite{gurukar2025long,di2025streaming,yang2025streammem}. However, continuous streaming is fundamentally an end-to-end throughput problem: the system must process video faster than it arrives. Accelerating one stage does not necessarily improve overall throughput, as bottlenecks may simply shift elsewhere. For example, \Cref{fig:streaming_pipelines} shows that token dropping reduces LLM computation but cannot prevent latency from growing with stream length, eventually limiting sustainable ingestion speed. This motivates our central idea: \emph{treating streaming VideoLLM inference as a coordinated multi-stage pipeline}, where bottlenecks can be identified and bounded to jointly optimize video-ingestion throughput and query-time TTFT.

To this end, we propose \textbf{ViCoStream} (\textbf{Vi}deo \textbf{Co}ordinated \textbf{Stream}ing), a stage-wise coordinated streaming framework that organizes visual preprocessing, visual encoding, token dropping, and LLM inference into a pipelined architecture where different video chunks progress through different thread-based stages concurrently. At the visual front end, incoming frames are accumulated into chunks and processed through overlapped CUDA-stream preprocessing and chunk-wise ViT encoding, improving GPU utilization and exploiting batch parallelism over frame-by-frame execution~\cite{li2025videoscan,chen2024videollm}. The projector then performs chunk-level temporal token dropping to regulate visual token density, reducing the computation and memory demands of downstream LLM processing~\cite{huang2025prunevid,chen2025streamingtom,wu2026hidrop}. Finally, the LLM employs bounded visual attention together with query-side retrieval, restricting the amount of visible visual history while preserving access to relevant past observations~\cite{zhang2024flash,zhang2026hermes,zhang2024sparsevlm}. As a result, the per-chunk prefill cost remains bounded and no longer grows with stream length, enabling arbitrarily long video streams under fixed latency and memory budgets. Collectively, these coordinated controls transform continuous streaming into a slowest-stage scheduling problem and decouple query-time TTFT from streaming FPS. Experiments show that ViCoStream achieves 134 FPS under 30\% token retention while keeping most moderate-compression settings on par with full-history baselines.

Our main contributions are as follows:
\begin{enumerate}
\setlength{\itemsep}{-0.2em}

\item We formulate continuous VideoLLM inference as a stage-wise streaming pipeline and identify two primary system objectives: sustaining video-ingestion throughput and minimizing query-time TTFT. This perspective exposes how bottlenecks migrate across stages and provides a unified framework for analyzing streaming efficiency.

\item We propose \textbf{ViCoStream}, a pipelined architecture that jointly coordinates visual processing and LLM inference. By combining stage-level parallelism with bounded visual-context management, our design maintains fixed per-chunk latency and memory costs independent of total stream length.

\item We provide a systematic efficiency--accuracy study of streaming VideoLLMs, analyzing how chunk granularity, token retention, visual-context size, and retrieval scope affect bottleneck migration and overall system performance. Our approach achieves up to 134 FPS while maintaining competitive accuracy.
\end{enumerate}

\section{Parallel Streaming Pipeline}
\label{sec:streaming_pipelines}

We formalize streaming VideoLLM inference as a stage-wise parallel pipeline. This formulation specifies how video data enters the system, how it is processed across stages, and how overall streaming throughput is determined under a shared resource model.

\paragraph{Chunk-wise Streaming Input}
A video stream is represented as a sequence of frames
\[
V=\{f_1,f_2,\dots,f_T\},
\]
which arrive sequentially over time. Since future frames are unavailable in streaming settings, the model must continuously process observed frames while maintaining a state for future query answering under bounded latency constraints.

To enable efficient processing, we partition the stream into non-overlapping chunks of size $c$. The $t$-th chunk is defined as
\[
\mathbf{V}_t = \{ f_{(t-1)c+1}, \dots, f_{tc} \}.
\]
Setting $c=1$ corresponds to frame-by-frame processing. In practice, chunking is beneficial because ViT encoders achieve higher efficiency through batched processing compared to serial per-frame execution.

\paragraph{Pipeline Components}
For each incoming chunk, streaming VideoLLM inference consists of four sequential stages.

\emph{Visual preprocessing} decodes raw frames and performs resizing, normalization, and sampling.

\emph{Visual encoding} maps the preprocessed frames into visual features using a ViT encoder:
\[
\mathbf{X}^{(v)}_t = E_v(\mathbf{V}_t).
\]

\emph{Token dropping} projects visual features into the language embedding space and removes redundant tokens:
\[
\mathbf{Z}_t = P(\mathbf{X}^{(v)}_t),
\]
where $\mathbf{Z}_t$ denotes the retained visual tokens.

\emph{LLM inference} incorporates visual tokens into the language model context, performs prefilling for user queries over the accumulated visual history, and generates autoregressive outputs.

\paragraph{Parallel Streaming Throughput}
While a single chunk traverses these stages sequentially, a streaming system can execute different chunks concurrently across stages. We adopt a stage-wise resource model in which each stage processes at most one chunk at a time. Concurrency arises from pipelining: different stages operate on different chunks simultaneously after pipeline warm-up.

Under this model, at time $t$, different chunks may reside in different stages (e.g., visual preprocessing, ViT encoding, token dropping and LLM inference). Let the per-chunk processing time be denoted by $T_{\mathrm{vp}}$, $T_{\mathrm{vit}}$, $T_{\mathrm{drop}}$, and $T_{\mathrm{llm}}$, respectively. Note that $T_{\mathrm{llm}}$ may vary with the amount of visible visual context. In steady state, the pipeline throughput is governed by the slowest stage:
\[
T_{\mathrm{pipe}} = \max \{T_{\mathrm{vp}}, T_{\mathrm{vit}}, T_{\mathrm{drop}}, T_{\mathrm{llm}}\}.
\]

The corresponding maximum streaming frame rate is:
\[
\mathrm{FPS}_{\max} = \frac{1000 \cdot c}{T_{\mathrm{pipe}}},
\]
where time is measured in milliseconds.

\paragraph{Streaming Latency Metrics}
Streaming performance has two complementary aspects. The first is continuous video ingestion throughput, captured by $\mathrm{FPS}_{\max}$, which measures how quickly the system can process incoming frames. The second is query-time latency, measured by time-to-first-token (TTFT), which includes LLM prefilling and the generation of the first output token after a complete user query is submitted over the observed video context. TTFT is measured from the moment when the complete user query is submitted to the LLM, using the video state observed up to the query timestamp. Frames arriving afterward may continue through the background pipeline, but are not used as evidence for the current response.

In this work, we use $\mathrm{FPS}_{\max}$ to characterize streaming throughput, while evaluating TTFT separately as the user-facing response latency.

\begin{figure*}[t]
    \centering
    \includegraphics[width=\textwidth]{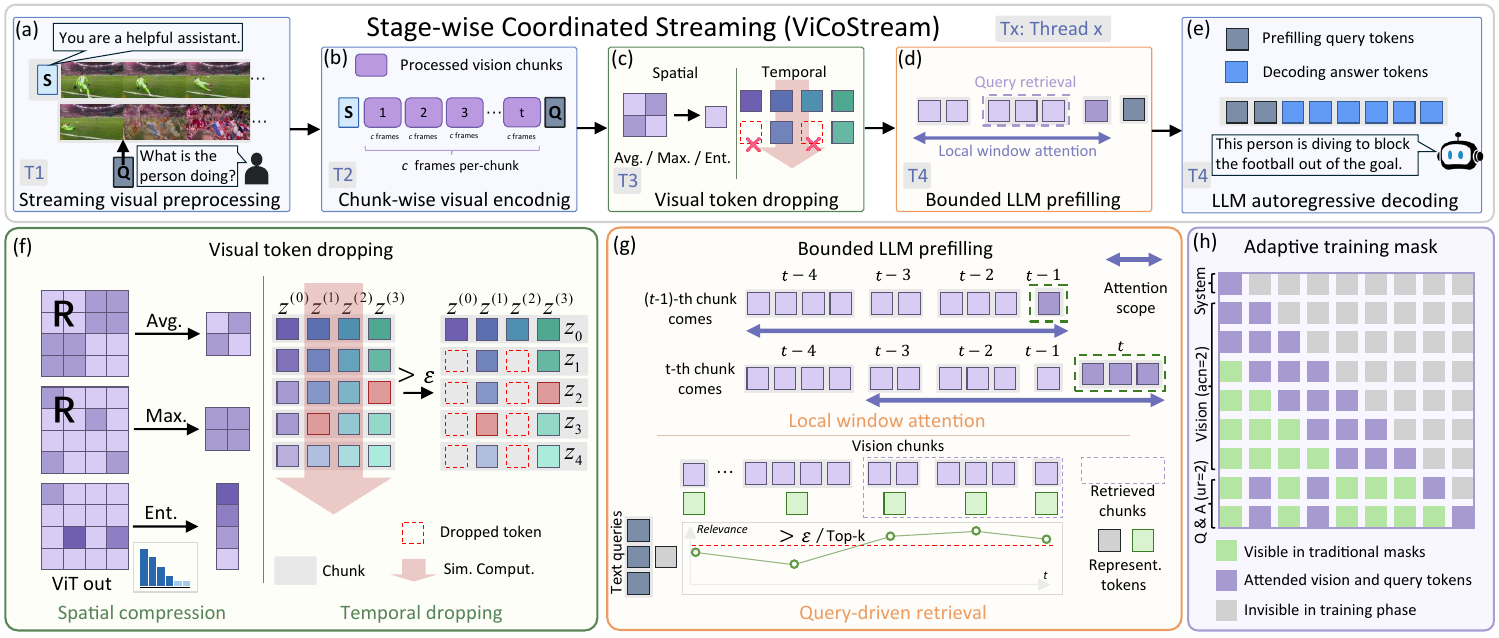}
    \caption{Overview of stage-wise coordinated streaming (ViCoStream). (a)--(e) show the main parallel pipeline from streaming input to autoregressive decoding in different threads. (d) and (e) share a thread due to their dependencies. (f) is spatial and temporal visual token compression, (g) is vision-based locality and query-driven retrieval for cross-attention scope, and (h) shows the adaptive training mask that matches these vision and query attention.}
    \label{fig:overview}
    \vspace{-0.5cm}
\end{figure*}

\section{Stage-Wise Coordinated Streaming}
\label{sec:coordinated_design}

The pipeline formulation in \Cref{sec:streaming_pipelines} suggests that bounded real-time streaming requires controlling each stage rather than only reducing a single cost term. We therefore organize coordinated streaming around four inference stages: visual preprocessing, visual encoding, token dropping, and LLM prefilling/decoding. In addition, we use an adaptive attention mask to align training-time visibility with the bounded vision-side attention and query-side retrieval used at inference. For each inference stage, we identify the corresponding bottleneck and choose a lightweight control that keeps the continuous visual-ingestion pipeline bounded, as summarized in \Cref{fig:overview}.

\paragraph{Visual Preprocessing}
Visual preprocessing samples, resizes, and normalizes incoming frames. Although this stage is not part of the neural model, it can become the throughput ceiling once LLM-side cost is reduced. We therefore treat visual preprocessing as the first stage of the streaming pipeline. When preprocessing involves GPU-side operations such as tensor resizing, normalization, patchification, or host-to-device transfer, we assign them to an independent CUDA stream so that this stage can overlap with later model stages rather than being serialized on the default stream. In steady state, visual preprocessing can prepare chunk $t{+}3$ while visual encoding, token dropping, and LLM prefilling process earlier chunks. This does not remove preprocessing cost, but it prevents preprocessing latency from being unnecessarily serialized with the later stages.

\paragraph{Visual Encoding}
Visual encoding determines how quickly new visual tokens are produced. Frame-level encoding invokes the visual encoder frequently and introduces high scheduling overhead, while overly large chunks delay token availability and increase per-chunk latency. We therefore use chunk-wise visual encoding, where consecutive frames are grouped into chunks and encoded together. The latency breakdown in \Cref{tab:latency} also shows why this granularity matters: increasing $c$ changes both VP and ViT latency and shifts the maximum streaming FPS, rather than simply scaling all stages uniformly.

\begin{figure*}[t]
    \centering
    \begin{minipage}[t]{0.26\textwidth}
        \centering
        \vspace{-3.05cm}
        \resizebox{\linewidth}{!}{\setlength{\tabcolsep}{3pt}
\fontsize{8.5pt}{9pt}\selectfont
\renewcommand{\arraystretch}{1.05}
\begin{tabular}{lcc}
\toprule
\textbf{Context} & \textbf{Drop} & \textbf{All} \\
\midrule
Full tokens & 0.0\% & 70.76 \\
Nearest-8K & 49.6\% & 71.68 \\
Nearest-4K & 64.5\% & \textbf{72.96} \\
Nearest-2K & 73.4\% & 71.76 \\
Nearest-1K & 82.2\% & 71.40 \\
\bottomrule
\end{tabular}
}
        \captionof{table}{Nearest visual context results  with Qwen2.5-VL-3B under full attention.}
        \label{tab:nearest_context_motivation}
    \end{minipage}
    \hfill
    \begin{minipage}[t]{0.34\textwidth}
        \centering
        \includegraphics[width=\linewidth]{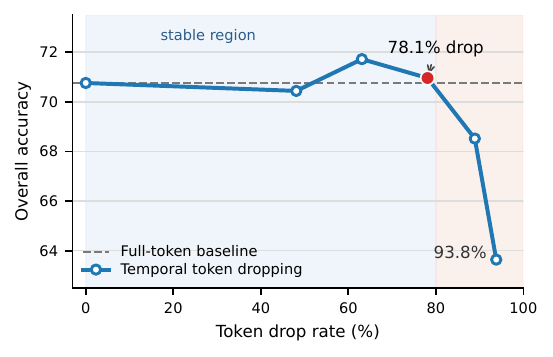}
        \vspace{-0.8cm}
        \captionof{figure}{Token drop rate versus accuracy on StreamingBench with trained Qwen2.5-VL-3B under full attention.}
        \label{fig:token_redundancy_motivation}
    \end{minipage}
    \hfill
    \begin{minipage}[t]{0.34\textwidth}
        \centering
        \includegraphics[width=\linewidth]{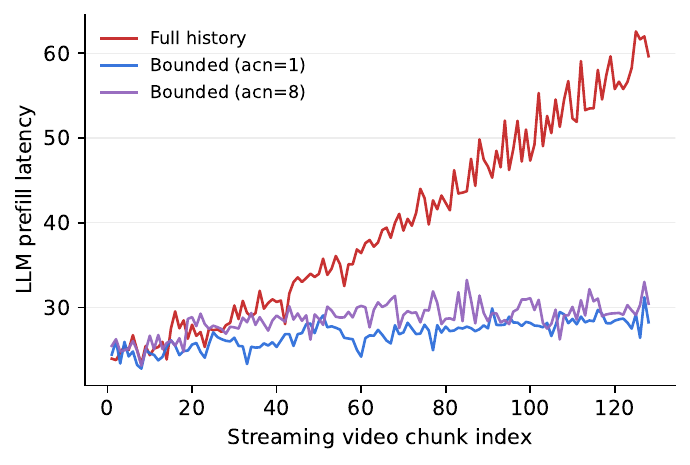}
        \vspace{-0.8cm}
        \captionof{figure}{LLM prefill latency over long streams. Full-history attention grows, while bounded attention remains stable.}
        \label{fig:latency_scaling}
    \end{minipage}
    \vspace{-0.25cm}
\end{figure*}

\paragraph{Token Dropping}
The projector and token-dropping stage determines how many visual tokens are passed to the LLM. Dense visual token sequences are highly redundant in streaming videos~\cite{yao2025timechat}. We observe the same pattern under full attention: as shown in \Cref{fig:token_redundancy_motivation}, dropping 78.1\% of visual tokens preserves the overall accuracy relative to the full-token baseline, while accuracy only drops substantially under more aggressive compression. This motivates token dropping as a direct way to reduce LLM prefilling cost before attention is applied.

We consider both spatial and temporal token dropping, as illustrated in \Cref{fig:overview}(f). Spatial compression operates at the frame level. Given projected visual tokens $X_v^{t} \in \mathbb{R}^{n_f \times d}$ for the $t$-th frame, where $n_f$ denotes the number of tokens per frame, we obtain a compact representation $\bar{X}_v^{t} \in \mathbb{R}^{\bar{n}_f \times d}$ with $\bar{n}_f < n_f$. This operation is applied independently to each frame, regardless of whether frames are processed individually or grouped into chunks. We evaluate three spatial strategies. 
\emph{Average pooling} aggregates tokens in a non-overlapping $2\times2$ region $\mathcal{R}_i$ as $\bar{x}_i=\frac{1}{|\mathcal{R}_i|}\sum_{x_j\in\mathcal{R}_i}x_j$, preserving region-level semantics at reduced spatial resolution.
\emph{Max pooling} retains the strongest activation in each region, $\bar{x}_i=\max_{x_j\in\mathcal{R}_i}x_j$, and discards less salient tokens.
\emph{Entropy-based selection} measures token importance by feature entropy $\mathcal{I}(x_j)=-\sum_m p_{jm}\log p_{jm}$, where $p_{jm}$ denotes the normalized probability of the $m$-th feature dimension, and retains the top-$K$ tokens with the highest entropy scores.

Temporal compression is applied at the chunk level and we always preserve the first frame of each chunk as an anchor, so every time segment leaves a residual visual record even when most pixels remain unchanged. For each spatial position $p$, we compute pairwise cosine similarity between the corresponding projected tokens in every frame pair $i,j\in\{1,\dots,c\}$ with $i<j$:
\begin{equation}
\mathcal{S}_{t,i \to j}(p)=
\frac{\langle \mathbf{z}_t^{(i)}(p),\mathbf{z}_t^{(j)}(p)\rangle}
{\|\mathbf{z}_t^{(i)}(p)\|\,\|\mathbf{z}_t^{(j)}(p)\|}.
\end{equation}

Here, $\mathbf{z}_t^{(i)}(p)\in\mathbb{R}^{d}$ denotes the projected feature of the token at spatial position $p$ in the $i$-th frame of the $t$-th chunk. A token at position $p$ is dropped when all pairwise similarities exceed a threshold $\tau$:
\begin{equation}
\mathrm{drop}_t(p)\Longleftrightarrow
\bigwedge_{i<j}\left[\mathcal{S}_{t,i\to j}(p)>\tau\right].
\end{equation}

This rule removes only tokens that are consistently redundant across the whole chunk. The preserved first-frame anchor provides a stable reference for what occurred in the segment and reduces the risk of error accumulation under aggressive temporal pruning.

When $c{=}1$, this formulation degenerates into frame-level temporal dropping, where each frame is treated as an independent dropping unit. In this setting, we prune tokens according to the projector-level similarity between the current frame and the previous frame, using the same threshold-based rule but without a multi-frame chunk anchor.

\paragraph{LLM Prefilling and Decoding}
The LLM stage consumes retained visual tokens through prefilling and later responds to user queries. Full-history attention makes this stage grow with accumulated visual context, so the key bottleneck is the historical visual cache visible to each incoming chunk.

Our empirical observations support a locality assumption for this stage: under full attention, retaining the nearest 4K visual tokens improves the overall accuracy from 70.76 to 72.96 while dropping 64.5\% of tokens, as shown in \Cref{tab:nearest_context_motivation}. This suggests that each incoming vision chunk often relies on nearby visual context rather than the entire accumulated history. As illustrated in \Cref{fig:overview}(g), we therefore adopt two complementary attention scopes: vision-side local prefilling and query-driven retrieval. For vision-side prefilling, we restrict each incoming chunk to attend only to the most recent $\textit{acn}$ chunks. Let $\mathbf{Q}_v^{(t)}$, $\mathbf{K}_v^{(t)}$, and $\mathbf{V}_v^{(t)}$ denote the query, key, and value projections of the current visual chunk. The attended keys and values are $\hat{\mathbf{K}}_v^{(t)}=\{\mathbf{K}_v^{(t-\textit{acn})},\dots,\mathbf{K}_v^{(t)}\}$ and $\hat{\mathbf{V}}_v^{(t)}=\{\mathbf{V}_v^{(t-\textit{acn})},\dots,\mathbf{V}_v^{(t)}\}$. The bounded visual attention is then computed as $\mathrm{Attn}_v^{(t)}=\mathrm{softmax}(\mathbf{Q}_v^{(t)}(\hat{\mathbf{K}}_v^{(t)})^\top/\sqrt{d})\hat{\mathbf{V}}_v^{(t)}$. This bounds the visible cache size and prevents LLM prefilling from growing with stream length.

For query-time decoding, the user query can introduce a different access pattern because relevant evidence may appear at non-recent positions in the stream. To bound query-time visual access without forcing full-history attention, we retrieve a limited set of visual chunks according to query--chunk similarity. The query tokens $\mathbf{X}_q$ and each visual chunk $\mathbf{X}_v^{(i)}$ are averaged into representative features $\bar{\mathbf{x}}_q$ and $\bar{\mathbf{x}}_v^{(i)}$, and their cosine similarity is
\begin{equation}
s_i=
\frac{\langle \bar{\mathbf{x}}_q,\bar{\mathbf{x}}_v^{(i)}\rangle}
{\|\bar{\mathbf{x}}_q\|\,\|\bar{\mathbf{x}}_v^{(i)}\|}.
\label{eq:query_retrieval}
\end{equation}

Only the top-$k$ chunks are exposed to query tokens during decoding. Let $\tilde{\mathbf{K}}_v^{(t)}$ and $\tilde{\mathbf{V}}_v^{(t)}$ denote the key and value caches constructed from these retrieved chunks. The query-side attention is computed as $\mathrm{Attn}_q^{(t)}=\mathrm{softmax}(\mathbf{Q}_q(\tilde{\mathbf{K}}_v^{(t)})^\top/\sqrt{d})\tilde{\mathbf{V}}_v^{(t)}$. Together, token dropping, bounded visual prefilling, and query-side retrieval convert the LLM cost from a history-dependent term into a bounded per-chunk and bounded per-query cost.

To align training with these two LLM attention scopes, we construct an additive attention mask $M\in\{0,-\infty\}^{T\times T}$, as illustrated in \Cref{fig:overview}(h). The mask assigns different visibility patterns to vision tokens and query tokens. For a vision token in the $t$-th chunk, attention is restricted to the most recent $\textit{acn}$ chunks, giving the visible set $\mathcal{V}_{\mathrm{vis}}(t)=\{j\mid t-\textit{acn}\leq j\leq t\}$. This produces a staircase-shaped local band in the vision--vision region of the mask. For a query token, visibility is determined by query-driven retrieval: only chunks whose similarity score $s_j$ exceeds a threshold $\tau_q$ remain visible, giving $\mathcal{V}_{\mathrm{qry}}=\{j\mid s_j\geq\tau_q\}$. Thus the query--vision region can be sparse and non-contiguous, allowing query tokens to access semantically relevant chunks regardless of temporal distance. Formally,
\begin{equation}
M_{i,j}=
\begin{cases}
 0, & t_j\in\mathcal{V}_{\mathrm{vis}}(t_i),\ i\in\mathcal{I}_{\mathrm{vis}},\\
 0, & t_j\in\mathcal{V}_{\mathrm{qry}},\ i\in\mathcal{I}_{\mathrm{qry}},\\
 -\infty, & \text{otherwise},
\end{cases}
\label{eq:adaptive_attention_mask}
\end{equation}
where $0$ leaves an attention logit unchanged and $-\infty$ masks it before softmax, $\mathcal{I}_{\mathrm{vis}}$ and $\mathcal{I}_{\mathrm{qry}}$ denote the indices of vision and query tokens, respectively, and $t_i$ and $t_j$ denote the chunk indices associated with tokens $i$ and $j$. This unified mask enforces local visual prefilling while preserving query-specific access to retrieved visual evidence.

\section{Experiments}

\subsection{Experimental Setup}
\paragraph{Datasets}
We train our models on two video-language instruction datasets: LLaVA-Video-100K~\cite{zhang2024llava} and TimeChat-Online-139K~\cite{yao2025timechat}. Unless otherwise specified, models are trained on their union. More dataset preprocessing details are provided in Appendix~\ref{app:data_preprocessing}.

We evaluate models on six streaming-oriented benchmarks: StreamingBench~\cite{lin2026streamingbench}, OvO-Bench~\cite{li2501ovo}, StreamBench~\cite{xiong2025streaming}, VStream-Ego~\cite{zhang2024flash}, VStream-Movie~\cite{zhang2024flash}, and ETBench~\cite{qian2025dispider}. These benchmarks cover real-time visual understanding, online video question answering, long-duration stream comprehension, and event-time reasoning. For each benchmark, we follow its official evaluation protocol and report the corresponding accuracy or task-level score; ETBench is summarized separately because it mixes heterogeneous metrics across sub-tasks. For StreamBench and VStream, we use GPT-3.5-Turbo as the judge for the score metric following the setup of Flash-VStream~\cite{zhang2024flash}.
In compact tables, we use SBench to denote StreamingBench.

\paragraph{Implementation Details}
Experiments are mainly conducted on Qwen2.5-VL-3B-Instruct and Qwen2.5-VL-7B-Instruct. Unless otherwise specified, our ablation studies use 3B-level models trained on a 50\% random subset of each dataset. During fine-tuning, the ViT encoder is frozen while the multimodal aligner and language model are updated. Models are trained on up to four H100 GPUs, while all evaluation and latency measurements are performed on a single A100 GPU. The ViCoStream implementation details are summarized in Appendix~\ref{app:coordinated_implementation}.

\subsection{Efficiency and Accuracy Analysis}
We first evaluate the end-to-end behavior of ViCoStream after jointly applying the stage-wise controls introduced above. This Section focuses on the resulting stage-wise latency, bottleneck migration, long-duration stability, and accuracy retention under real-time streaming constraints. We defer component-level analyses of visual preprocessing, visual encoding, token dropping, intra-chunk attention scope, attention locality, and query retrieval to the Section \ref{sec:ablation_studies}.

\begin{table}[t]
\centering
\footnotesize
\setlength{\tabcolsep}{6pt}
\renewcommand{\arraystretch}{1.1}

\setlength{\aboverulesep}{0pt}
\setlength{\belowrulesep}{0pt}
\begin{tabular}{c c c c c c}
\toprule
$c$ & \textbf{VP} & \textbf{ViT} & \textbf{Drop} & \textbf{LLM Prefill} & \textbf{FPS} \\
\midrule
\rowcolor[gray]{0.92}
\multicolumn{6}{c}{\textbf{Token Retention 30\%} \enspace ($r{\approx}70\%$)} \\
1  & 5   & 30  & 6  & \underline{38} & 26 \\
4  & 37  & 40  & 12 & \underline{44} & 91 \\
8  & \underline{73}  & 64  & 24 & 44 & 110 \\
16 & 113 & \underline{119} & 44 & 42 & \textbf{134} \\
32 & \underline{248} & 235 & 87 & 69 & 129 \\
\midrule
\rowcolor[gray]{0.92}
\multicolumn{6}{c}{\textbf{Token Retention 50\%} \enspace ($r{\approx}50\%$)} \\
1  & 5   & 30  & 6  & \underline{38}  & 26 \\
4  & \underline{47}  & 41  & 12 & 45  & 86 \\
8  & \underline{82}  & 64  & 23 & 45  & 98 \\
16 & \underline{160} & 119 & 45 & 64  & 100 \\
32 & \underline{317} & 236 & 87 & 124 & \textbf{101} \\
\bottomrule
\end{tabular}

\caption{Per-module latency breakdown and maximum streaming FPS under ViCoStream ($\textit{acn}{=}4$).
VP, ViT, Drop, and LLM Prefill denote visual preprocessing, visual encoding, token dropping, and LLM prefilling, respectively.
The underlined value marks the throughput bottleneck among the streaming stages.}
\label{tab:latency}
\vspace{-0.5cm}
\end{table}

\begin{table*}[t]
\centering
\footnotesize
\setlength{\tabcolsep}{3.4pt}
\renewcommand{\arraystretch}{1.08}
\def\g#1{$_{\textcolor{blue!60!black}{\downarrow #1}}$}
\def\r#1{$_{\textcolor{red}{\downarrow #1}}$}
\def\p#1{$_{\textcolor{green!50!black}{\uparrow #1}}$}

\setlength{\aboverulesep}{0pt}
\setlength{\belowrulesep}{0pt}
\begin{tabular}{l c c c c c c c c}
\toprule
\textbf{Model} & $\mathit{acn}$ & $\bm{\textit{ur}}$ & $\mathit{s}$ & \textbf{StreamingBench} & \textbf{OvO-Bench} & \textbf{StreamBench} & \textbf{RVS-E} & \textbf{RVS-M} \\
\midrule
\rowcolor[gray]{0.92}
\multicolumn{9}{c}{\textbf{Traditional Streaming Prefill (Baseline)}} \\
Qwen2.5-VL-3B & $\infty$ & $\infty$ & -- & 70.32 & 47.13 & 48.37 & 34.59 & 33.91 \\
Qwen2.5-VL-7B & $\infty$ & $\infty$ & -- & 74.60 & 49.35 & 54.90 & 34.86 & 34.60 \\
\midrule
\rowcolor[gray]{0.92}
\multicolumn{9}{c}{\textbf{ViCoStream\textsuperscript{\textdagger}}} \\
\multirow{8}{*}{Qwen2.5-VL-3B\textsuperscript{\textdagger}} & 4 & 64 & 0.25 ($\downarrow$57.6\%) & 70.44\p{0.1} & 45.30\g{1.8} & 49.72\p{1.4} & 35.22\p{0.6} & 31.97\g{1.9} \\
& 8 & 64 & 0.25 ($\downarrow$57.3\%) & 70.52\p{0.2} & 46.35\g{0.8} & 49.06\p{0.7} & 34.12\g{0.5} & 31.55\g{2.4} \\
& 4 & 64 & 0.1 ($\downarrow$72.9\%) & 68.32\g{2.0} & 42.19\r{4.9} & 46.35\g{2.0} & 34.40\g{0.2} & 29.04\r{4.9} \\
& 8 & 64 & 0.1 ($\downarrow$73.0\%) & 67.39\g{2.9} & 43.28\r{3.9} & 47.52\g{0.9} & 35.43\p{0.8} & 32.25\g{1.7} \\
& 4 & 32 & 0.25 ($\downarrow$57.6\%) & 70.68\p{0.4} & 45.16\g{2.0} & 49.94\p{1.6} & 32.29\g{2.3} & 31.98\g{1.9} \\
& 8 & 32 & 0.25 ($\downarrow$57.3\%) & 70.48\p{0.2} & 46.17\g{1.0} & 49.14\p{0.8} & 32.18\g{2.4} & 30.93\g{3.0} \\
& 4 & 32 & 0.1 ($\downarrow$72.9\%) & 68.28\g{2.0} & 42.48\r{4.7} & 46.09\g{2.3} & 30.99\r{3.6} & 28.39\r{5.5} \\
& 8 & 32 & 0.1 ($\downarrow$72.9\%) & 67.68\g{2.6} & 43.85\r{3.3} & 47.66\g{0.7} & 32.70\g{1.9} & 31.34\g{2.6} \\

\cmidrule{1-9}
\multirow{8}{*}{Qwen2.5-VL-7B\textsuperscript{\textdagger}} & 4 & 64 & 0.25 ($\downarrow$48.0\%) & 71.64\g{3.0} & 49.41\p{0.1} & 54.07\g{0.8} & 36.68\p{1.8} & 31.67\g{2.9} \\
& 8 & 64 & 0.25 ($\downarrow$48.4\%) & 71.80\g{2.8} & 50.77\p{1.4} & 53.68\g{1.2} & 34.31\g{0.6} & 33.23\g{1.4} \\
& 4 & 64 & 0.1 ($\downarrow$70.6\%) & 69.71\r{4.9} & 47.56\g{1.8} & 51.01\r{3.9} & 32.70\g{2.2} & 28.14\r{6.5} \\
& 8 & 64 & 0.1 ($\downarrow$70.5\%) & 70.42\r{4.2} & 47.23\g{2.1} & 50.97\r{3.9} & 35.63\p{0.8} & 31.07\r{3.5} \\
& 4 & 32 & 0.25 ($\downarrow$48.0\%) & 71.40\r{3.2} & 49.71\p{0.4} & 53.51\g{1.4} & 36.71\p{1.9} & 31.70\g{2.9} \\
& 8 & 32 & 0.25 ($\downarrow$48.4\%) & 71.48\r{3.1} & 50.35\p{1.0} & 53.51\g{1.4} & 34.40\g{0.5} & 33.02\g{1.6} \\
& 4 & 32 & 0.1 ($\downarrow$70.6\%) & 69.56\r{5.0} & 46.77\g{2.6} & 50.53\r{4.4} & 32.12\g{2.7} & 28.30\r{6.3} \\
& 8 & 32 & 0.1 ($\downarrow$70.5\%) & 70.44\r{4.2} & 47.32\g{2.0} & 50.42\r{4.5} & 34.88\p{0.0} & 31.76\g{2.8} \\
\bottomrule
\end{tabular}

\caption{Accuracy under ViCoStream.
\textbf{Baseline} uses full cross-attention over all accumulated visual tokens without compression.
\textsuperscript{\textdagger} denotes ViCoStream with temporal token compression and bounded cross-attention. Token drop rates in parentheses are measured on StreamingBench, and subscripts indicate \textcolor{green!50!black}{gains}, \textcolor{blue!60!black}{<3\% drops}, or \textcolor{red}{>3\% drops} relative to baseline. Values are percentages.}
\label{tab:main_accuracy}
\end{table*}

\begin{table*}[tbph]
\centering
\scriptsize
\setlength{\tabcolsep}{2.5pt}
\renewcommand{\arraystretch}{1.1}
\def\g#1{$_{\textcolor{blue!60!black}{\downarrow #1}}$}
\def\r#1{$_{\textcolor{red}{\downarrow #1}}$}
\def\p#1{$_{\textcolor{green!50!black}{\uparrow #1}}$}
\resizebox{\textwidth}{!}{%
\begin{tabular}{l c c c c c c c c c c c c}
\toprule
\textbf{Model} & $\mathit{acn}$ & $\bm{\textit{ur}}$ & $\mathit{s}$ & \textbf{ECA} $\uparrow$ & \textbf{EPM} (mIoU$\uparrow$ / F1$\uparrow$) & \textbf{GVQ} $\uparrow$ & \textbf{RAR} $\uparrow$ & \textbf{RVQ} $\uparrow$ & \textbf{TAL-PT} $\uparrow$ & \textbf{TAL-TH} $\uparrow$ & \textbf{TEM} $\uparrow$ & \textbf{TVG} $\uparrow$ \\
\midrule
Qwen2.5-VL-3B & $\infty$ & $\infty$ & -- & 37.80 & 1.29 / 1.60 & 53.45 & 34.80 & 51.60 & 18.63 & 13.09 & 16.80 & 27.30 \\
Qwen2.5-VL-7B & $\infty$ & $\infty$ & -- & 45.00 & 1.66 / 2.10 & 56.55 & 36.60 & 54.60 & 20.17 & 12.61 & 26.70 & 33.35 \\
\midrule
\multirow{2}{*}{Qwen2.5-VL-3B\textsuperscript{\textdagger}} & 4 & 32 & 0.25 & 37.20\g{0.6} & 1.54\p{0.2} / 1.75\p{0.1} & 52.07\g{1.4} & 35.80\p{1.0} & 48.20\r{3.4} & 21.74\p{3.1} & 14.66\p{1.6} & 20.40\p{3.6} & 32.10\p{4.8} \\
& 4 & 32 & 0.1 & 37.20\g{0.6} & 1.62\p{0.3} / 1.95\p{0.3} & 54.48\p{1.0} & 35.20\p{0.4} & 45.60\r{6.0} & 21.63\p{3.0} & 14.42\p{1.3} & 16.35\g{0.4} & 30.35\p{3.1} \\
\cmidrule{1-13}
\multirow{2}{*}{Qwen2.5-VL-7B\textsuperscript{\textdagger}} & 4 & 32 & 0.25 & 37.60\r{7.4} & 2.04\p{0.4} / 2.50\p{0.4} & 54.48\g{2.1} & 35.60\g{1.0} & 51.20\r{3.4} & 17.95\g{2.2} & 12.51\g{0.1} & 21.80\r{4.9} & 28.30\r{5.1} \\
& 4 & 32 & 0.1 & 37.00\r{8.0} & 1.25\g{0.4} / 1.50\g{0.6} & 53.10\r{3.4} & 33.60\g{3.0} & 46.60\r{8.0} & 12.22\r{8.0} & 12.66\p{0.1} & 21.40\r{5.3} & 26.40\r{7.0} \\
\bottomrule
\end{tabular}%
}

\caption{ETBench task-level results. ECA, GVQ, RAR, and RVQ are accuracy-based tasks; EPM is temporal grounding with mIoU/F1; TAL-PT, TAL-TH, and TVG are F1-based temporal localization tasks; TEM reports mean recall. \textsuperscript{\textdagger} denotes ViCoStream with temporal token compression and local cross-attention. Values are percentages.}
\label{tab:etbench}
\vspace{-0.5cm}
\end{table*}

\paragraph{Stage-Wise Latency and Maximum FPS}
\Cref{tab:latency} reports the per-module latency breakdown under different streaming configurations according to \Cref{sec:streaming_pipelines}. All latency values are measured in milliseconds on a single A100 GPU with simulated 720p input resized to $448{\times}448$. The low throughput at $c{=}1$ shows why frame-level processing is inefficient: frequent small ViT invocations and model calls limit the stream to 26 FPS. Chunk-wise processing amortizes this overhead and raises the maximum streaming FPS to 134 under 30\% token retention and 101 under 50\% token retention, although larger chunks are not always better because VP, ViT, and token-dropping latency also increase with $c$. Under coordinated parallel streaming, each stage processes one chunk at a time, but different stages can process different chunks concurrently; thus throughput is determined by the slowest continuous stage rather than the sum of all stages (i.e., $\mathrm{FPS}_{\max}=1000c/T_{\mathrm{stage}}$). With bounded attention (e.g., $\textit{acn}{=}4$), LLM prefilling remains bounded instead of growing with full history, shifting the limiting stage from LLM prefilling at small chunks to ViT encoding or preprocessing-related throughput at larger chunks. Higher token retention further lowers the attainable FPS by increasing LLM prefilling cost. These results show that coordinated parallel streaming supports high-frame-rate real-time ingestion while revealing bottleneck migration: once LLM-side cost is reduced, visual encoding and preprocessing become the next throughput ceilings.

\paragraph{Long-Duration Latency Stability}
We next examine whether LLM prefilling remains bounded as visual history accumulates.
As shown in \Cref{fig:latency_scaling}, full-history prefilling makes each incoming chunk attend to an increasingly long visual cache, so LLM prefill latency rises steadily with the streaming chunk index.
By contrast, bounded attention with $\textit{acn}{=}1$ or $\textit{acn}{=}8$ keeps the visible visual context fixed and yields nearly flat per-chunk latency.
This confirms that attention locality converts LLM prefilling from a history-dependent cost into a bounded per-chunk cost, which is necessary for long-duration streaming input.

\paragraph{Query-Side Response Latency}
\Cref{fig:query_response_latency} (Left) further shows that TTFT is governed by the product of retrieval scope and token retention. When only 25\% or 50\% of tokens are kept, increasing the query-visible chunks has a modest latency impact. However, at 75\% or 100\% retention, TTFT rises sharply once the query attends to 16 or more chunks. Thus, query-side retrieval should be tuned jointly with token dropping: exposing more chunks is only affordable when each chunk has already been sufficiently compressed. Therefore, the decoding stage will not become a new bottleneck under sufficiently aggressive token retention.
\Cref{fig:streaming_ttft_query_selection} (Middle) further compares full-history query access with bounded query selection across 25\% token retention. With $\textit{ur}{=}32$ or $\textit{ur}{=}64$, query-side prefill latency is reduced to below 50 ms relative to full-history access, indicating that retrieval keeps the query-prefill cost nearly constant even as the accumulated stream grows.
In our measured configurations, even a conservative serial sum of visual preprocessing, visual encoding, token dropping, and vision-side prefilling for the last observed chunk is at most 764 ms in \Cref{tab:latency}, below typical spoken-query duration. Thus, the reported TTFT isolates query-side prefilling and first-token generation, while chunk-processing cost is captured by the stage-wise latency analysis.

\begin{table}[t]
\centering
\fontsize{8.5pt}{9.pt}\selectfont

\setlength{\tabcolsep}{3.2pt}
\renewcommand{\arraystretch}{1.05}

\setlength{\aboverulesep}{0pt}
\setlength{\belowrulesep}{0pt}
\begin{tabular}{c c c c c c}
\toprule
\multirow{2}{*}{\textbf{$s$}} & \multirow{2}{*}{\textbf{Full Attn.}} & \multirow{2}{*}{\textbf{StreamingBench}} & \multicolumn{3}{c}{\textbf{OvO-Bench}} \\
\cmidrule(lr){4-6}
& & & \textbf{Bwd.} & \textbf{Real.} & \textbf{Fwd.} \\
\midrule
\rowcolor[gray]{0.92}
\multicolumn{6}{c}{\textbf{Qwen2.5-VL-3B}} \\
\multirow{2}{*}{0.25} & \ding{51}  & 69.96 & 39.33 & 54.53 & 34.96 \\
                      & \ding{55} & 70.44 & 37.36 & 55.74 & 32.35 \\
\cmidrule(lr){1-6}
\multirow{2}{*}{0.1}  & \ding{51}  & 67.40 & 40.72 & 52.13 & 34.80 \\
                      & \ding{55} & 68.32 & 34.78 & 50.45 & 33.07 \\
\midrule
\rowcolor[gray]{0.92}
\multicolumn{6}{c}{\textbf{Qwen2.5-VL-7B}} \\
\multirow{2}{*}{0.25} & \ding{51}  & 73.28 & 44.14 & 59.71 & 39.14 \\
                      & \ding{55} & 71.64 & 42.44 & 60.11 & 35.01 \\
\cmidrule(lr){1-6}
\multirow{2}{*}{0.1}  & \ding{51}  & 71.36 & 43.54 & 55.51 & 37.65 \\
                      & \ding{55} & 69.71 & 42.15 & 57.70 & 32.69 \\
\bottomrule
\end{tabular}

\caption{Comparison of full and bounded intra-chunk prefill under ViCoStream.
}
\label{tab:ablation_intrachunk}
\end{table}

\begin{table}[t]
\centering
\footnotesize
\setlength{\tabcolsep}{5pt}
\renewcommand{\arraystretch}{1.05}
\begin{tabular}{c c c c}
\toprule
$c$ & \textbf{Traditional VP} & \textbf{CUDA-stream VP} & \textbf{Speedup} \\
\midrule
1  & 52.96   & 4.48   & 11.8$\times$ \\
4  & 227.59  & 29.38  & 7.7$\times$ \\
8  & 461.40  & 59.49  & 7.8$\times$ \\
16 & 989.48  & 122.22 & 8.1$\times$ \\
32 & 1961.17 & 244.25 & 8.0$\times$ \\
\bottomrule
\end{tabular}
\caption{Visual preprocessing latency (ms) under traditional preprocessing and CUDA-stream preprocessing.}
\label{tab:vp_profiling}
\vspace{-0.5cm}
\end{table}

\paragraph{Accuracy Preservation Under ViCoStream}
After analyzing stage-wise latency, we evaluate the full ViCoStream configuration by jointly applying token dropping, bounded visual attention, and query-side retrieval. The goal is not to claim a standalone accuracy improvement, but to measure how much accuracy is retained once the pipeline is constrained for real-time streaming. For each base model, we compare against traditional streaming prefill with full accumulated visual context and no compression ($r=0\%$, $\textit{acn}=\infty$, $\textit{ur}=\infty$), where $\textit{ur}$ denotes the number of visual chunks retrieved or attended by the user query. This within-model comparison isolates the accuracy cost introduced by the coordinated streaming controls.

\Cref{tab:main_accuracy} shows that moderate token dropping provides a favorable trade-off: with $s{=}0.25$, most results remain within 3 points of the full-history baseline for both 3B and 7B models, and several settings improve on StreamingBench, StreamBench, or RVS-E. This suggests that full visual history is often unnecessary once recent visual attention and query-side retrieval are preserved. More aggressive dropping ($s{=}0.1$) exposes the failure modes: StreamingBench and StreamBench remain relatively robust, whereas OvO-Bench and RVS-M degrade more often, indicating higher sensitivity to fine temporal evidence or movie-level context. 

\Cref{tab:etbench} further breaks this trend down by event-time tasks under $\textit{ur}{=}32$. For the 3B model, temporal localization metrics such as TAL-PT, TAL-TH, and TVG improve despite bounded context, suggesting that local attention can remove redundant history while preserving event evidence. The 7B model is less stable on ETBench, with larger drops on ECA, TEM, and TVG, indicating stronger sensitivity to compressed visual history in some event-time tasks.

The comparisons above establish the accuracy competitiveness of stage-wise coordinated streaming, but they do not factor out token dropping as a controlled variable: the baseline uses full visual context without compression, while the coordinated variants combine dropping with bounded attention and query-side retrieval. To isolate the effect of attention scope from the effect of token reduction, we compare full-attention and bounded-attention intra-chunk prefill under the same chunk-level token-dropping configuration. As shown in \Cref{tab:ablation_intrachunk}, bounded attention can slightly help StreamingBench and real-time perception for the 3B model, but it can weaken memory-heavy OvO-Bench splits such as backward tracing under stronger dropping. Overall, neither attention scope is uniformly better across models and tasks. The accuracy trade-off is therefore dominated by the token-dropping and retrieval controls, while the intra-chunk attention scope mainly changes how temporal evidence is exposed to the LLM during coordinated parallel execution.

\subsection{Ablation Studies}
\label{sec:ablation_studies}

The preceding analysis establishes that ViCoStream achieves real-time video inference with competitive accuracy. We now examine each pipeline stage individually to understand per-component bottlenecks and the resulting accuracy--latency trade-offs. The front-end ablations analyze the VP and ViT stages that limit throughput once LLM-side cost is reduced, while the token dropping, attention locality, and retrieval ablations characterize the controls that govern the remaining efficiency--accuracy balance.
\begin{table}[htbp]
\centering
\footnotesize
\setlength{\tabcolsep}{10pt}
\renewcommand{\arraystretch}{1.0}
\begin{tabular}{l c c c c}
\toprule
\textbf{Input Setting} & $c$ & $\mathit{acn}$ & $\mathit{s}$ & \textbf{All} \\
\midrule
Baseline & -- & $\infty$ & -- & 70.32 \\
\midrule
\multirow{2}{*}{Frame-level} & 1 & 16 & 0.25 & 68.80 \\
& 1 & 16 & 0.1 & 66.48 \\
\midrule
\multirow{4}{*}{Chunk-level} & 4 & 4 & 0.25 & 70.68 \\
& 4 & 4 & 0.1 & 68.28 \\
& 4 & 8 & 0.25 & 70.48 \\
& 4 & 8 & 0.1 & 67.68 \\
\bottomrule
\end{tabular}
\caption{Comparison of frame-level and chunk-level temporal token dropping on StreamingBench. The appendix reports the full per-task breakdown.}
\label{tab:frame_level_dropping_summary}
\vspace{-0.5cm}
\end{table}

\begin{table*}[htbp]
\centering
\begin{minipage}[t]{0.34\textwidth}
\centering
\scriptsize
\setlength{\tabcolsep}{2pt}
\renewcommand{\arraystretch}{1.08}
\begin{tabular*}{\linewidth}{@{\extracolsep{\fill}}l c c c c}
\toprule
\multirow{2}{*}{\textbf{Threshold} ($\mathit{s}$)} & \multirow{2}{*}{\textbf{SBench}} & \multicolumn{3}{c}{\textbf{OvO-Bench}} \\
\cmidrule(lr){3-5}
& & \textbf{Bwd.} & \textbf{Real.} & \textbf{Fwd.} \\
\midrule
Baseline ($s{=}1.0$, 0\%) & 70.32 & 41.04 & 56.72 & 34.04 \\
$s{=}0.5$ ($\downarrow$19\%) & 68.48 & 39.06 & 57.52 & 32.88 \\
$s{=}0.25$ ($\downarrow$60\%) & 67.64 & 37.33 & 55.25 & 33.42 \\
$s{=}0.2$ ($\downarrow$67\%) & 67.68 & 37.51 & 54.90 & 33.01 \\
$s{=}0.15$ ($\downarrow$71\%) & 66.76 & 37.20 & 53.49 & 32.28 \\
$s{=}0.1$ ($\downarrow$74\%) & 66.56 & 38.08 & 52.07 & 32.22 \\
\bottomrule
\end{tabular*}
\end{minipage}
\hfill
\begin{minipage}[t]{0.34\textwidth}
\centering
\scriptsize
\setlength{\tabcolsep}{2pt}
\renewcommand{\arraystretch}{1.25}
\begin{tabular*}{\linewidth}{@{\extracolsep{\fill}}l c c c c}
\toprule
\multirow{2}{*}{\textbf{Spatial Strat.}} & \multirow{2}{*}{\textbf{SBench}} & \multicolumn{3}{c}{\textbf{OvO-Bench}} \\
\cmidrule(lr){3-5}
& & \textbf{Bwd.} & \textbf{Real.} & \textbf{Fwd.} \\
\midrule
Baseline & 70.32 & 41.04 & 56.72 & 34.04 \\
Avg. ($\downarrow$71.3\%) & 66.60 & 36.41 & 53.43 & 31.64 \\
Max. ($\downarrow$71.3\%) & 65.99 & 37.15 & 52.13 & 32.51 \\
Ent. ($\downarrow$50.2\%) & 67.63 & 40.62 & 53.62 & 32.66 \\
Ent. ($\downarrow$74.7\%) & 66.51 & 35.69 & 53.22 & 32.85 \\
\bottomrule
\end{tabular*}
\end{minipage}
\hfill
\begin{minipage}[t]{0.28\textwidth}
\centering
\scriptsize
\setlength{\tabcolsep}{2pt}
\renewcommand{\arraystretch}{1.08}
\setlength{\tabcolsep}{1.5pt}
\begin{tabular*}{\linewidth}{@{\extracolsep{\fill}}l c c c c}
\toprule
\multirow{2}{*}{\textbf{Compression}} & \multirow{2}{*}{\textbf{SBench}} & \multicolumn{3}{c}{\textbf{OvO-Bench}} \\
\cmidrule(lr){3-5}
& & \textbf{Bwd.} & \textbf{Real.} & \textbf{Fwd.} \\
\midrule
None & 70.32 & 41.04 & 56.72 & 34.04 \\
S$_{\mathrm{Ent.}}$ & 67.63 & 40.62 & 53.62 & 32.66 \\
S$_{\mathrm{Avg.}}$ & 66.60 & 36.41 & 53.43 & 31.64 \\
T$_{s=0.25}$ & 70.44 & 37.36 & 55.74 & 32.35 \\
T$_{s=0.1}$ & 68.32 & 34.78 & 50.45 & 33.07 \\
S$_{\mathrm{Ent.}}$+T$_{s=0.25}$ & 65.95 & 36.07 & 54.44 & 31.99 \\
\bottomrule
\end{tabular*}
\end{minipage}
\caption{(Left) Temporal token compression under incremental streaming. (Middle) Spatial token compression strategies under the same setting. (Right) Component-wise ablation combining spatial and temporal compression.}
\label{tab:ablation_temporal}
\label{tab:ablation_spatial}
\label{tab:ablation_component}
\vspace{-0.25cm}
\end{table*}

\begin{figure*}[htbp]
\centering
\begin{minipage}[t]{0.32\textwidth}
\vspace{0pt}
\centering
\includegraphics[width=\linewidth]{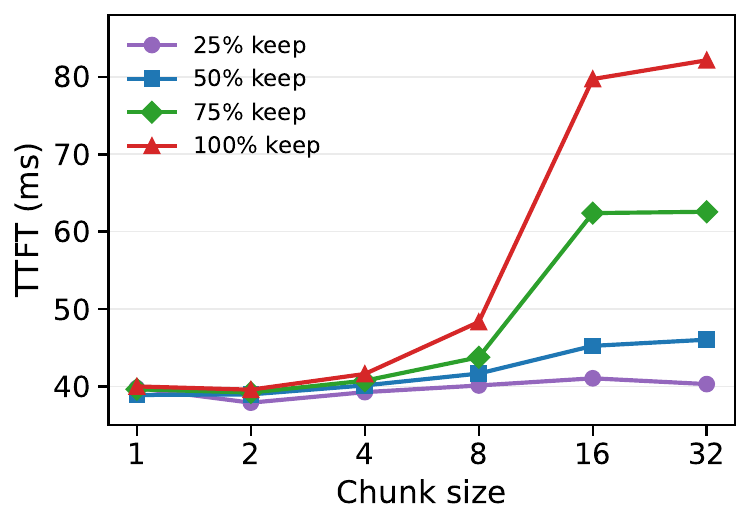}
\end{minipage}
\hfill
\begin{minipage}[t]{0.32\textwidth}
\vspace{0pt}
\centering
\includegraphics[width=\linewidth]{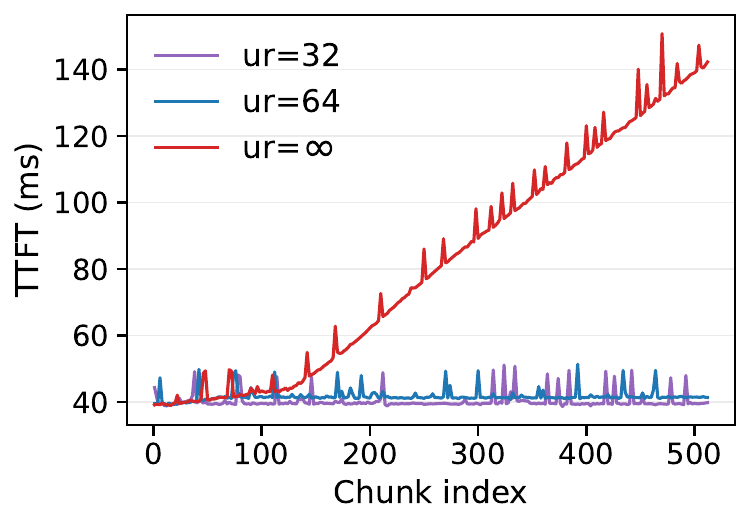}
\end{minipage}
\hfill
\begin{minipage}[t]{0.32\textwidth}
\vspace{0pt}
\centering
\includegraphics[width=\linewidth]{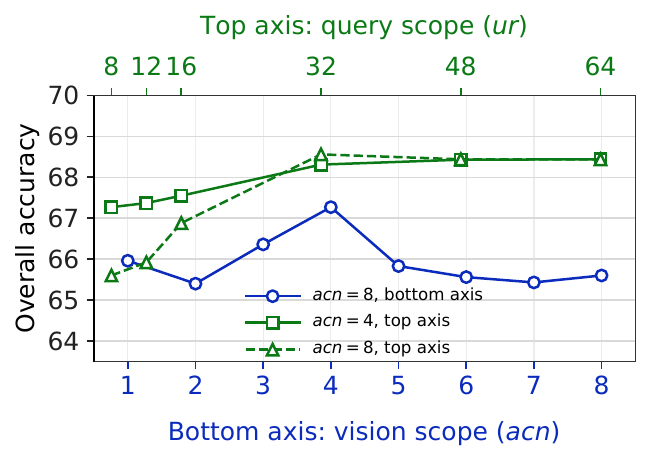}
\end{minipage}
\vspace{-0.25cm}

\caption{Latency and accuracy under bounded cross-attention. (Left) Query response latency (TTFT) under different token retention and retrieval budgets. (Middle) Query-side prefill TTFT under full-history access and bounded query-selection budgets. (Right) Accuracy under different vision-side and query-side attention scopes.}
\label{fig:query_response_latency}
\label{fig:streaming_ttft_query_selection}
\label{fig:attention_locality}
\vspace{-0.25cm}
\end{figure*}

\paragraph{Visual Preprocessing}
Visual preprocessing mainly determines whether incoming frames can be prepared fast enough for the downstream model stages. \Cref{tab:vp_profiling} shows that CUDA-stream VP substantially reduces preprocessing latency compared with traditional CPU-based VP, with roughly $8\times$ speedup for larger chunks. Locally, this removes the CPU preprocessing bottleneck for each chunk, but globally the optimized VP cost still grows with chunk size, so preprocessing remains a real throughput stage once LLM-side cost is bounded.

\paragraph{Visual Encoding}
Visual encoding is governed by the ViT stage, where chunk-wise processing is the main alternative to frame-level encoding. As reflected in \Cref{tab:latency}, increasing the chunk size amortizes frequent small ViT invocations, but it also raises the per-chunk ViT latency. This explains why throughput improves from frame-level processing to moderate chunk sizes, while larger chunks can shift the limiting stage from LLM prefilling to visual encoding or preprocessing.

\paragraph{Token Dropping Ablations}
Before comparing token-compression strategies, we first examine the granularity of temporal token dropping. As defined in \Cref{sec:coordinated_design}, setting $c{=}1$ degenerates the chunk-level rule into frame-level temporal dropping, where each frame is compressed independently based on projector-level similarity to the previous frame. \Cref{tab:frame_level_dropping_summary} shows that frame-level dropping remains competitive at moderate thresholds, but becomes less robust under aggressive dropping. This supports our use of chunk-level temporal dropping, where chunk anchors provide a more stable unit for preserving temporal evidence.

With chunk-level granularity fixed, we next vary the strength of temporal reduction at fixed $c{=}4$, $\textit{acn}{=}4$, and $\textit{ur}{=}32$. \Cref{tab:ablation_temporal} (Left) shows that temporal dropping produces a relatively smooth trade-off: StreamingBench changes mildly across thresholds, while OvO-Bench is more task-dependent, with the backward-tracing and real-time perception splits generally more sensitive than the forward active-response. Moderate settings such as $s{=}0.25$ preserve most baseline performance, and even $s{=}0.1$ does not lead to severe degradation.

Temporal reduction operates on the chunk axis, while we now test the orthogonal spatial axis. Table \ref{tab:ablation_spatial} (Middle) shows that spatial compression is more sensitive to the selection strategy: average and max pooling consistently hurt accuracy, whereas entropy-based selection is more robust because it preserves semantically salient tokens. Compared with the main temporal-dropping results in \Cref{tab:main_accuracy}, these drops suggest that spatial compression does not directly transfer to projector-level token reduction; removing spatial tokens can discard object or region-level evidence that the LLM still relies on.

To test whether the two reductions compose, Table \ref{tab:ablation_component} (Right) compares spatial-only, temporal-only, and combined compression. It shows that spatial and temporal reduction are complementary rather than interchangeable. Combining them gives the strongest compression but also accumulates the largest accuracy loss. Overall, the results suggest that token dropping should not be treated as a uniform reduction problem: temporal redundancy can be removed relatively safely, whereas spatial compression must preserve salient visual evidence, and combining both dimensions should be reserved for settings that prioritize efficiency over the remaining accuracy margin.

\paragraph{LLM Attention Locality}
Token dropping reduces tokens per chunk but does not prevent visible history from growing with stream length, so bounded attention is needed to control LLM prefilling over long-duration inputs. As shown in \Cref{fig:attention_locality} (Right), $\textit{acn}$ is not monotonic: increasing the vision-side window to $\textit{acn}{=}4$ improves accuracy, but further increasing it to 8 brings no additional gains and can slightly reduce performance. The best region is a moderate local window around $\textit{acn}{=}4$, suggesting that recent chunks provide useful evidence while older chunks add redundancy or noise. On the query side, increasing $\textit{ur}$ helps initially but largely saturates around $\textit{ur}{=}32$ (about 128 frames when $c{=}4$). Thus, query-time access is treated as a compact retrieval budget that preserves evidence while bounding response latency.

\section{Conclusion}
We study why real-time streaming remains difficult for VideoLLMs even with compression or memory modules. We formulate streaming inference as a stage-wise coordinated pipeline over visual preprocessing, visual encoding, token dropping, and LLM prefilling/decoding, where continuous ingestion is limited by the slowest stage and response latency is reflected by TTFT. This view reveals that reducing LLM-side cost alone is insufficient: once visual tokens are compressed or attention is bounded, the bottleneck can migrate to visual encoding, preprocessing, or query-time retrieval. Our results show that ViCoStream enables high-frame-rate streaming while preserving accuracy close to full-history baselines.

\section*{Limitations}

\paragraph{Resource-Constrained Parallelism}
Our coordinated parallel streaming analysis assumes a stage-wise resource model: each stage processes one chunk at a time, while different stages can process different chunks concurrently. This makes bottleneck analysis tractable and matches pipeline-style streaming deployments, but it may not fully capture systems with multiple replicas per stage, heterogeneous accelerators, or more aggressive asynchronous scheduling. Future work can extend the framework to richer hardware settings and study bottlenecks under flexible resource allocation.

\paragraph{Accuracy--Latency Trade-Off}
Decomposing streaming inference into a multi-stage pipeline helps measure latency and memory bottlenecks in true streaming tasks, but it does not remove the accuracy--latency trade-off. Token dropping, bounded attention, and query-side retrieval improve efficiency while often preserving accuracy, yet some tasks remain sensitive to compressed or bounded visual history. Future work should explore adaptive policies that allocate context according to task difficulty, query intent, and stream dynamics.

\bibliography{ref}

\clearpage
\appendix

\section{Related Work}
\paragraph{Streaming Video Understanding}
Streaming video understanding requires models to process continuously arriving frames and remain responsive to user queries.
Many long-video VideoLLMs improve efficiency by sampling frames~\cite{li2024videochat}, compressing visual tokens~\cite{bolya2022token,shang2025llava,xing2024pyramiddrop,shen2024longvu,wang2025adaretake,wu2026data,wu2026hidrop}, or retrieving query-relevant visual context~\cite{zhang2024sparsevlm,huang2025prunevid} from a complete video context, but such designs often assume that the full clip or user query is available before compression and are therefore closer to offline processing than strict real-time streaming.

Recent streaming systems process visual inputs incrementally and optimize different stages of the pipeline. For visual preprocessing and visual encoding, TimeChat-Online~\cite{yao2025timechat}, VideoScan~\cite{li2025videoscan}, VideoLLM-MoD~\cite{wu2024videollm}, and CodecSight~\cite{zou2026codecsight} reduce incoming visual redundancy, route vision computation, or exploit codec signals before dense model inference. For token dropping, StreamingTOM~\cite{chen2025streamingtom} reduces redundant temporal observations during online processing, STC~\cite{wang2025stc} maintains a compact streaming token representation to limit LLM input growth, and KFFocus~\cite{nie2025kffocus} highlights keyframes so that more computation is allocated to informative visual moments. For LLM prefilling and decoding, StreamMem~\cite{yang2025streammem} summarizes long visual histories into memory states, Flash-VStream~\cite{zhang2024flash} uses compact memory to support efficient video streaming, and HERMES~\cite{zhang2026hermes} reduces the burden of long-context KV caching. Other online assistants focus on when to invoke or respond with the LLM: StreamMind~\cite{ding2025streammind} uses event-gated cognition to avoid unnecessary reasoning, Dispider~\cite{qian2025dispider} separates perception, decision, and reaction for timely interaction, Think-as-You-See~\cite{zhang2026think} explores streaming chain-of-thought reasoning, and LION-FS~\cite{li2025lion} combines fast and slow response paths for online video assistance. StreamOV~\cite{xie2026streamov} further extends streaming video understanding to omni-modal audio-visual interaction with evidence-guided long-short term memory and hidden-state response triggering. These component-level designs make online dialogue more practical, but they mainly address local stages of the streaming pipeline in isolation. In contrast, our work analyzes streaming video inference from a multi-stage coordinated streaming perspective, studying how stage-wise controls jointly affect efficiency bottlenecks.

\section{Experimental Details}

\subsection{Dataset Preprocessing}
\label{app:data_preprocessing}
TimeChat-Online-139K is directly adopted from TimeChat-Online~\cite{yao2025timechat}. For LLaVA-Video-100K~\cite{zhang2024llava}, we use the video list provided by TimeChat-Online. During preprocessing, videos are sampled at 1 FPS. For video sequences longer than 64 frames, we apply adaptive-stride uniform sampling during training, so that the retained frames cover a sufficiently broad temporal range.

\subsection{Evaluation Benchmarks}
\label{app:evaluation_benchmarks}
We evaluate coordinated streaming on five benchmark sources that cover complementary aspects of streaming video understanding. \textbf{StreamingBench}~\cite{lin2026streamingbench} evaluates streaming video understanding through time-dependent questions asked along the video timeline; we use its Real-time Visual Understanding subset, which focuses on recognizing objects, actions, scene changes, and other visual evidence available at the current streaming moment. \textbf{OVO-Bench}~\cite{li2501ovo} evaluates online video understanding with timestamped queries under backward tracing, real-time perception, and forward active-response settings, allowing us to test whether bounded context affects different temporal reasoning modes. \textbf{StreamBench}~\cite{xiong2025streaming} is designed for streaming video understanding and multi-round interaction, and is used to assess long-duration online comprehension with memory-intensive queries. \textbf{VStream-QA}~\cite{zhang2024flash} provides timestamped question-answer pairs over ego-centric and movie videos; we report results on VStream-Ego and VStream-Movie to separate online egocentric understanding from long-form narrative reasoning. \textbf{ETBench}~\cite{qian2025dispider} contains event-level and time-sensitive video-language tasks. We use the six ETBench subsets most suitable for streaming video understanding: Charades-STA, Ego4D-NLQ, QA-Ego4D, STAR, Perception Test, and THUMOS14. These subsets involve timestamped event localization, action recognition, or question answering grounded in specific temporal evidence, so they naturally test whether a model can preserve event-level information while processing videos incrementally.

\paragraph{Baseline protocol}
Our primary comparisons use the same Qwen2.5-VL backbone with traditional full-history streaming prefill as the reference setting. This within-backbone protocol is chosen because the paper studies how coordinated streaming controls change latency, memory, and accuracy under a fixed model, rather than comparing different VideoLLM systems. Existing streaming VideoLLMs often differ in backbone, instruction data, visual preprocessing, prompt format, memory design, and whether they assume delayed or truly incremental input; directly comparing their final accuracy would therefore mix the effect of our pipeline with many uncontrolled factors. Moreover, many prior systems do not report the stage-wise latency, continuous-ingestion FPS, TTFT, and peak-memory measurements needed for our bottleneck-migration analysis.

\subsection{Coordinated Streaming Implementation}
\label{app:coordinated_implementation}

\paragraph{Training setup}
For the coordinated streaming variants, we fine-tune Qwen2.5-VL-Instruct backbones with full-parameter supervised fine-tuning on the union of TimeChat-Online-139K and LLaVA-Video-100K. The ViT encoder is frozen, while the multimodal aligner and language model are updated. Training uses bfloat16 precision with DeepSpeed ZeRO-2 on up to four GPUs. We set the per-device batch size to 1, gradient accumulation steps to 16, learning rate to $1\times10^{-5}$, warmup ratio to 0.05, maximum sequence length to 8192, and train for one epoch. Visual inputs are capped with a pixel budget of 90K pixels for both images and videos, corresponding to an approximately $300{\times}300$ resolution budget before model-specific resizing.

\subsection{Detailed Benchmark Results}

\paragraph{StreamBench and VStream}
\Cref{tab:streambench_vstream_detailed} reports accuracy and score on StreamBench, RVS-Ego, and RVS-Movie. StreamBench remains robust under coordinated streaming: moderate dropping improves the 3B model over the full-history baseline and only mildly reduces the 7B model. RVS-Ego also benefits from bounded processing in several settings, indicating that egocentric streaming questions often rely on recent or locally retrievable evidence. In contrast, RVS-Movie is consistently more fragile, with most coordinated settings below the full-history baseline, suggesting that movie-level understanding depends more heavily on long-range narrative context. These results complement the main table by showing that the same efficiency controls are most reliable for online visual understanding and more costly for long-form story-level reasoning.

\begin{table}[t]
\centering
\fontsize{7.pt}{8.pt}\selectfont

\setlength{\tabcolsep}{1.8pt}
\renewcommand{\arraystretch}{1.12}
\resizebox{\columnwidth}{!}{%
\setlength{\aboverulesep}{0pt}
\setlength{\belowrulesep}{0pt}
\begin{tabular}{l c c c cc cc cc}
\toprule
\multirow{2}{*}{\textbf{Model}} & \multirow{2}{*}{$\mathit{acn}$} & \multirow{2}{*}{$\bm{\textit{ur}}$} & \multirow{2}{*}{$\mathit{s}$} & \multicolumn{2}{c}{\textbf{StBench}} & \multicolumn{2}{c}{\textbf{RVS-E}} & \multicolumn{2}{c}{\textbf{RVS-M}} \\
\cmidrule(lr){5-6}\cmidrule(lr){7-8}\cmidrule(lr){9-10}
& & & & \textbf{Acc.} & \textbf{Sc.} & \textbf{Acc.} & \textbf{Sc.} & \textbf{Acc.} & \textbf{Sc.} \\
\midrule
\rowcolor[gray]{0.92}
\multicolumn{10}{c}{\textbf{Traditional Streaming Prefill (Baseline)}} \\
3B & $\infty$ & $\infty$ & -- & 48.37 & 3.07 & 34.59 & 3.21 & 33.91 & 2.95 \\
7B & $\infty$ & $\infty$ & -- & 54.90 & 3.22 & 34.86 & 3.21 & 34.60 & 2.98 \\
\midrule
\rowcolor[gray]{0.92}
\multicolumn{10}{c}{\textbf{ViCoStream\textsuperscript{\textdagger}}} \\
\multirow{8}{*}{3B\textsuperscript{\textdagger}} & 4 & 64 & 0.25 & 49.72 & 3.12 & 35.22 & 3.35 & 31.97 & 2.94 \\
& 8 & 64 & 0.25 & 49.06 & 3.10 & 34.12 & 3.24 & 31.55 & 2.90 \\
& 4 & 64 & 0.1 & 46.35 & 2.97 & 34.40 & 3.33 & 29.04 & 2.87 \\
& 8 & 64 & 0.1 & 47.52 & 3.00 & 35.43 & 3.29 & 32.25 & 2.91 \\
& 4 & 32 & 0.25 & 49.94 & 3.12 & 32.29 & 3.24 & 31.98 & 2.92 \\
& 8 & 32 & 0.25 & 49.14 & 3.11 & 32.18 & 3.13 & 30.93 & 2.89 \\
& 4 & 32 & 0.1 & 46.09 & 2.96 & 30.99 & 3.22 & 28.39 & 2.83 \\
& 8 & 32 & 0.1 & 47.66 & 3.00 & 32.70 & 3.28 & 31.34 & 2.89 \\
\cmidrule{1-10}
\multirow{8}{*}{7B\textsuperscript{\textdagger}} & 4 & 64 & 0.25 & 54.07 & 3.23 & 36.68 & 3.18 & 31.67 & 2.90 \\
& 8 & 64 & 0.25 & 53.68 & 3.22 & 34.31 & 3.14 & 33.23 & 2.97 \\
& 4 & 64 & 0.1 & 51.01 & 3.15 & 32.70 & 3.09 & 28.14 & 2.77 \\
& 8 & 64 & 0.1 & 50.97 & 3.13 & 35.63 & 3.17 & 31.07 & 2.89 \\
& 4 & 32 & 0.25 & 53.51 & 3.22 & 36.71 & 3.17 & 31.70 & 2.92 \\
& 8 & 32 & 0.25 & 53.51 & 3.22 & 34.40 & 3.16 & 33.02 & 2.95 \\
& 4 & 32 & 0.1 & 50.53 & 3.13 & 32.12 & 3.10 & 28.30 & 2.79 \\
& 8 & 32 & 0.1 & 50.42 & 3.13 & 34.88 & 3.15 & 31.76 & 2.90 \\
\bottomrule
\end{tabular}%
}
\caption{Expanded results on StreamBench (StBench), RVS-Ego (RVS-E), and RVS-Movie (RVS-M) using Qwen2.5-VL backbones in~\Cref{tab:main_accuracy}. Acc. and Sc. denote the official accuracy and score metrics of each benchmark judged by GPT-3.5-Turbo. \textsuperscript{\textdagger} denotes ViCoStream with token dropping and bounded attention.}
\label{tab:streambench_vstream_detailed}
\vspace{-0.5cm}
\end{table}

\paragraph{StreamingBench}

\Cref{tab:streamingbench_detailed} provides the per-task breakdown on StreamingBench, which focuses on real-time visual understanding, including object perception, causal reasoning, clip summarization, event understanding, and other online perception tasks. In this table, OP, CR, CS, ATP, EU, TR, PR, SU, ACP, and CT denote Object Perception, Causal Reasoning, Clips Summarization, Attribute Perception, Event Understanding, Text-Rich Understanding, Prospective Reasoning, Spatial Understanding, Action Perception, and Counting, respectively. The results show that coordinated streaming is especially stable under moderate temporal dropping ($s{=}0.25$): Qwen2.5-VL-3B slightly improves over the full-history baseline in overall accuracy, and several sub-tasks such as clip summarization, attribute perception, prospective reasoning, and action perception remain comparable or improve. The main degradation appears under stronger dropping ($s{=}0.1$), where counting and text-rich understanding become more sensitive. This suggests that StreamingBench contains substantial visual redundancy, but fine-grained counting and text-related cues still benefit from denser retained tokens.

\paragraph{OVO-Bench}

\Cref{tab:ovobench_detailed} breaks OvO-Bench into real-time visual perception, backward tracing, and forward active response. Real-time perception is relatively robust to bounded attention and token dropping: for the 7B model, several coordinated settings match or exceed the baseline average for this category. Backward tracing is more sensitive, especially for the 3B model under aggressive dropping, reflecting the need to recover evidence from earlier visual history. Forward active response changes less dramatically, partly because clue-revealing response remains stable across many settings. Overall, OvO-Bench confirms that the cost of coordinated streaming is task-dependent: online perception tolerates bounded context better than memory-heavy backward reasoning.

\section{Streaming Paradigms}
\label{app:streaming_paradigms}

\begin{figure*}[htbp]
    \centering
    \includegraphics[width=0.92\textwidth]{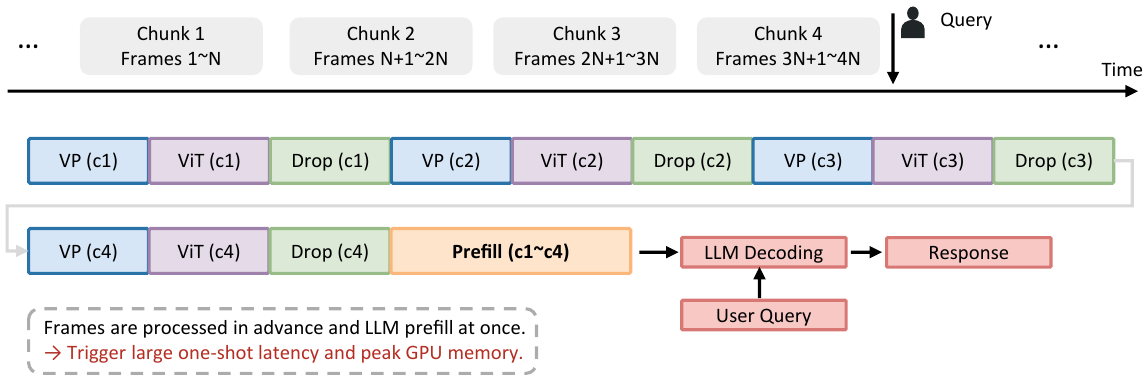}
    \vspace{1em}
    
    {\small\textbf{(a) Delayed streaming.} The full video is available before inference and can be cached or compressed in advance.}
    \vspace{1em}

    \includegraphics[width=0.92\textwidth]{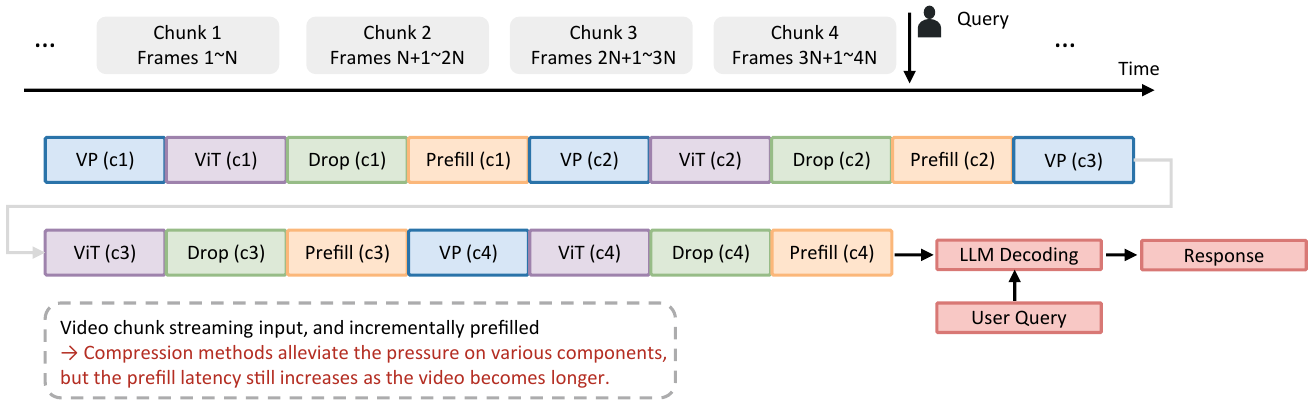}
    \vspace{1em}
    
    {\small\textbf{(b) Continuous streaming.} Video chunks arrive incrementally, but stages are executed sequentially for each chunk.}
    \vspace{1em}

    \includegraphics[width=0.92\textwidth]{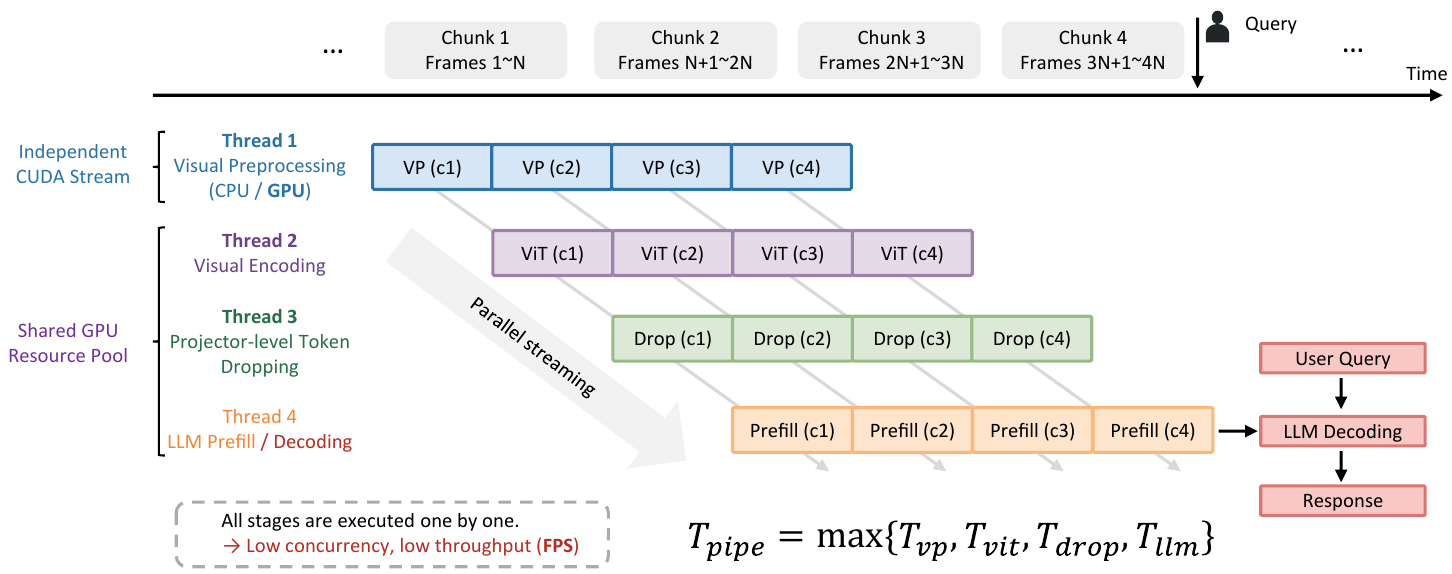}
    \vspace{1em}
    
    {\small\textbf{(c) Stage-wise coordinated streaming (Ours).} Different chunks occupy different stages in a pipeline, enabling stage-wise overlap.}
    \caption{Comparison of streaming inference paradigms for VideoLLMs.}
    \label{fig:streaming_paradigms}
\end{figure*}

\Cref{fig:streaming_paradigms}(a) illustrates \textbf{delayed streaming}, which corresponds to the fake-streaming setting mentioned in the Introduction. The complete video is pre-cached before inference, even if frames are originally delivered as a stream, and the cached context may be compressed before the user query arrives. The LLM performs a single one-shot prefill only at query time, analogous to a human watching the entire video first and only beginning to reason when asked a question. This paradigm has a light early-stage burden, since it mainly runs visual encoding when no LLM-side prefill is performed. However, if the cached context has not been sufficiently compressed, the query-time prefill concentrates a long visual history into one forward pass, producing large latency and GPU peak-memory spikes. It is therefore unsuitable for strict real-time streaming, where queries may arrive at arbitrary moments and responses must remain bounded.

\Cref{fig:streaming_paradigms}(b) illustrates \textbf{continuous streaming}, where video chunks are received incrementally and the LLM performs prefilling after each new chunk arrives. This is analogous to a human reasoning after each newly observed video segment before continuing to watch. Such systems can apply local acceleration techniques within individual components, such as token compression or memory reduction. However, since visual preprocessing, visual encoding, token dropping, and LLM prefilling are still executed serially for each chunk, the per-chunk latency may grow with the accumulated visual history. For long streams, the system may fail to finish all stages within one chunk interval, making long streaming video inference difficult.

\Cref{fig:streaming_paradigms}(c) illustrates our \textbf{stage-wise coordinated streaming}. It also receives video incrementally, but decomposes continuous inference into stage-wise pipeline slots: after warm-up, visual preprocessing, visual encoding, token dropping, and LLM prefilling operate on different chunks concurrently. The maximum supported streaming rate is therefore governed by $T_{\mathrm{pipe}}=\max\{T_{\mathrm{vp}},T_{\mathrm{vit}},T_{\mathrm{drop}},T_{\mathrm{llm}}\}$ rather than the sum of all stage latencies. This formulation exposes the true streaming bottleneck and reveals how it migrates after individual stages are optimized. In our experiments, stage-wise streaming supports over 130 FPS while keeping per-chunk latency and GPU peak memory bounded over long streams.

\section{Token-Dropping Visualization}
\begin{figure*}[t]
    \centering
    \includegraphics[width=0.9\textwidth]{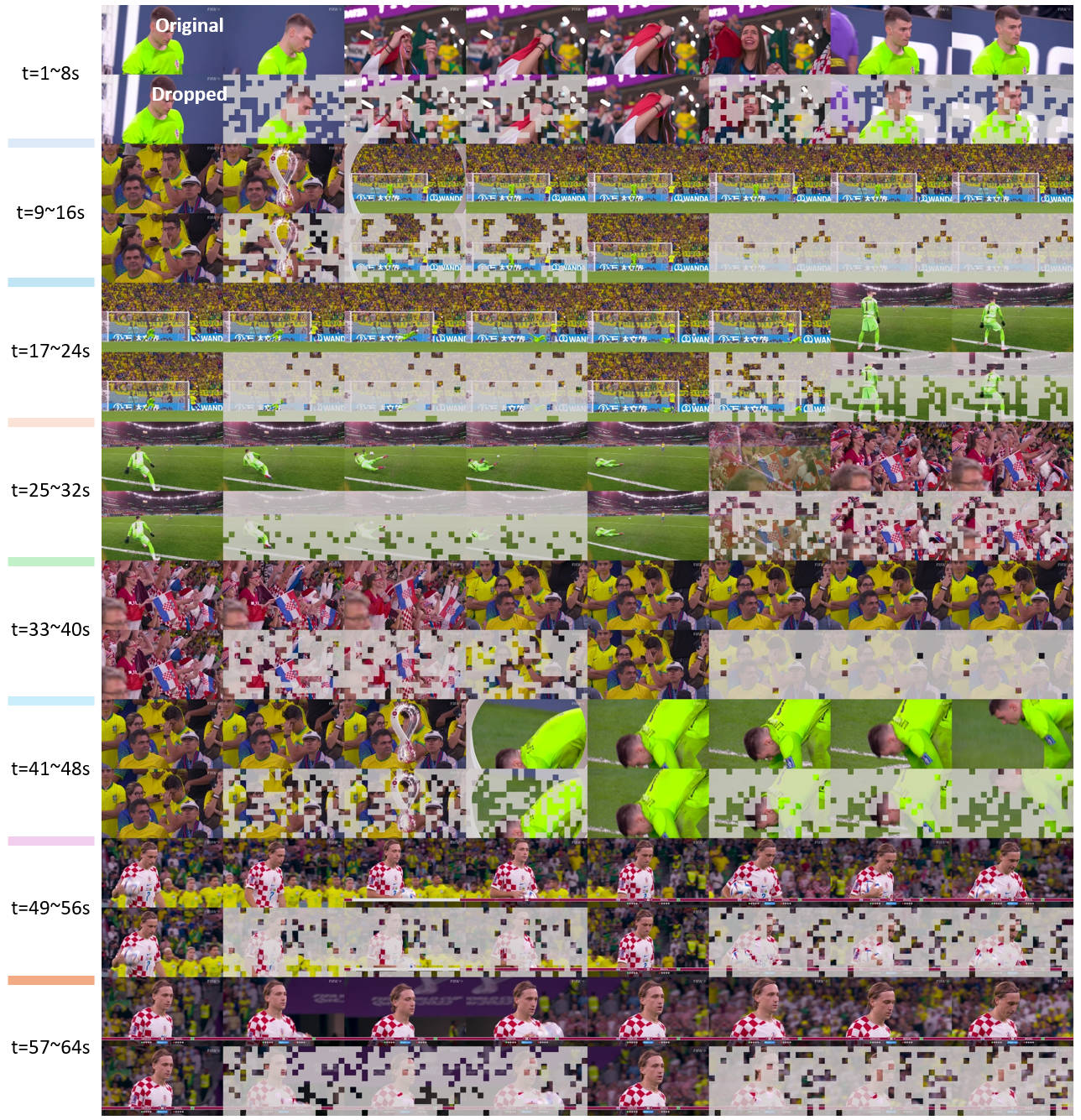}
    \vspace{-0.15cm}
    \caption{Visualization of intra-chunk temporal token dropping on StreamingBench. White patches indicate dropped visual tokens, while the remaining regions are retained for LLM prefilling.}
    \label{fig:token_drop_streamingbench}
    \vspace{-0.35cm}
\end{figure*}

\Cref{fig:token_drop_streamingbench} shows how the intra-chunk temporal token-dropping rule acts on a 64-second StreamingBench video sampled at 1 FPS. With chunk-wise processing and threshold $s{=}0.25$, tokens that remain redundant within each chunk are effectively removed, as indicated by the white patches. Meanwhile, the first frame of each chunk is kept as an anchor, preserving rich representative information about what occurs in that segment. This design maintains a high drop rate while reducing the risk of discarding all evidence for a temporally stable event.

\section{Frame-level Token Dropping}
\label{app:frame_level_dropping}

\Cref{tab:frame_level_dropping} reports the expanded frame-level temporal token dropping on StreamingBench, where every frame is treated as an independent dropping unit with chunk size $c{=}1$ and a per-frame similarity threshold $s$. The reported configurations use the same backbone as the main table, and the chunk-level rows are taken from \Cref{tab:streamingbench_detailed} for comparison.

The main text summarizes the key takeaway in \Cref{tab:frame_level_dropping_summary}. Here we keep the detailed per-task values corresponding to that summary, including the Counting-sensitive cases where the two granularities diverge most clearly.

\FloatBarrier

\begin{table*}[t]
\centering
\fontsize{6.8pt}{7.4pt}\selectfont

\setlength{\tabcolsep}{6.0pt}
\renewcommand{\arraystretch}{1.2}
\resizebox{\textwidth}{!}{%
\setlength{\aboverulesep}{0pt}
\setlength{\belowrulesep}{0pt}
\begin{tabular}{l c c c c c c c c c c c c c c}
\toprule
\textbf{Model} & $\mathit{acn}$ & $\bm{\textit{ur}}$ & $\mathit{s}$ & \textbf{OP} & \textbf{CR} & \textbf{CS} & \textbf{ATP} & \textbf{EU} & \textbf{TR} & \textbf{PR} & \textbf{SU} & \textbf{ACP} & \textbf{CT} & \textbf{All} \\
\midrule
\rowcolor[gray]{0.92}
\multicolumn{15}{c}{\textbf{Traditional Streaming Prefill (Baseline)}} \\
Qwen2.5-VL-3B & $\infty$ & $\infty$ & -- & 74.53 & 75.78 & 74.13 & 76.92 & 74.84 & 74.45 & 70.37 & 63.41 & 66.19 & 46.81 & 70.32 \\
Qwen2.5-VL-7B & $\infty$ & $\infty$ & -- & 80.49 & 83.59 & 78.86 & 81.41 & 77.36 & 78.19 & 72.22 & 67.48 & 69.32 & 50.53 & 74.60 \\
\midrule
\rowcolor[gray]{0.92}
\multicolumn{15}{c}{\textbf{ViCoStream\textsuperscript{\textdagger}}} \\
\multirow{8}{*}{Qwen2.5-VL-3B\textsuperscript{\textdagger}} & 4 & 64 & 0.25 & 73.98 & 70.31 & 75.71 & 77.56 & 71.07 & 73.83 & 71.30 & 63.41 & 69.60 & 46.81 & 70.44 \\
& 8 & 64 & 0.25 & 73.98 & 71.88 & 75.39 & 76.92 & 75.47 & 75.39 & 69.44 & 63.01 & 67.05 & 48.40 & 70.52 \\
& 4 & 64 & 0.1 & 70.19 & 71.88 & 75.39 & 75.64 & 72.96 & 69.16 & 69.44 & 62.60 & 67.90 & 40.43 & 68.32 \\
& 8 & 64 & 0.1 & 70.46 & 70.31 & 74.45 & 75.64 & 71.07 & 66.36 & 66.67 & 62.60 & 66.19 & 41.18 & 67.39 \\
& 4 & 32 & 0.25 & 73.17 & 69.53 & 75.39 & 77.88 & 70.44 & 74.45 & 72.22 & 63.41 & 69.60 & 51.06 & 70.68 \\
& 8 & 32 & 0.25 & 73.44 & 71.88 & 76.66 & 77.24 & 74.84 & 74.45 & 69.44 & 63.01 & 67.05 & 48.40 & 70.48 \\
& 4 & 32 & 0.1 & 69.92 & 70.31 & 74.45 & 76.28 & 72.96 & 69.78 & 70.37 & 62.20 & 68.47 & 39.89 & 68.28 \\
& 8 & 32 & 0.1 & 68.83 & 69.53 & 76.03 & 76.28 & 71.70 & 67.60 & 68.52 & 63.41 & 65.91 & 40.96 & 67.68 \\
\cmidrule{1-15}
\multirow{8}{*}{Qwen2.5-VL-7B\textsuperscript{\textdagger}} & 4 & 64 & 0.25 & 76.42 & 78.12 & 78.86 & 77.24 & 72.33 & 74.77 & 75.93 & 67.48 & 68.47 & 39.36 & 71.64 \\
& 8 & 64 & 0.25 & 75.88 & 78.91 & 79.50 & 77.24 & 77.36 & 74.77 & 76.85 & 66.26 & 68.75 & 37.23 & 71.80 \\
& 4 & 64 & 0.1 & 74.80 & 78.12 & 76.66 & 77.88 & 72.33 & 71.34 & 77.78 & 66.26 & 66.76 & 28.88 & 69.71 \\
& 8 & 64 & 0.1 & 75.27 & 77.34 & 77.29 & 77.17 & 73.42 & 70.72 & 74.07 & 65.85 & 67.24 & 38.42 & 70.42 \\
& 4 & 32 & 0.25 & 76.15 & 78.12 & 79.18 & 76.60 & 73.58 & 73.52 & 75.00 & 68.29 & 67.90 & 38.83 & 71.40 \\
& 8 & 32 & 0.25 & 76.42 & 78.91 & 80.13 & 75.96 & 75.47 & 73.83 & 76.85 & 66.67 & 67.61 & 37.77 & 71.48 \\
& 4 & 32 & 0.1 & 75.27 & 77.34 & 77.29 & 78.46 & 72.33 & 69.78 & 75.00 & 66.26 & 66.95 & 27.07 & 69.56 \\
& 8 & 32 & 0.1 & 74.72 & 77.34 & 77.29 & 77.17 & 74.36 & 71.34 & 74.04 & 66.12 & 68.19 & 36.81 & 70.44 \\
\bottomrule
\end{tabular}%
}
\caption{Expanded StreamingBench results on the Real-time Visual Understanding subset. }
\label{tab:streamingbench_detailed}
\vspace{-0.5cm}
\end{table*}

\begin{table*}[t]
\centering
\scriptsize
\setlength{\tabcolsep}{1.8pt}
\renewcommand{\arraystretch}{1.2}
\resizebox{\textwidth}{!}{%
\setlength{\aboverulesep}{0pt}
\setlength{\belowrulesep}{0pt}
\begin{tabular}{l c c c | c c c c c c c | c c c c | c c c c | c}
\toprule
\multirow{2}{*}{\textbf{Model}} & \multirow{2}{*}{$\mathit{acn}$} & \multirow{2}{*}{$\bm{\textit{ur}}$} & \multirow{2}{*}{$\mathit{s}$} & \multicolumn{7}{c|}{\textbf{Real-Time Visual Perception}} & \multicolumn{4}{c|}{\textbf{Backward Tracing}} & \multicolumn{4}{c|}{\textbf{Forward Active Response}} & \multirow{2}{*}{\textbf{All}} \\
\cmidrule(lr){5-11} \cmidrule(lr){12-15} \cmidrule(lr){16-19}
 & & & & \textbf{OCR} & \textbf{ACR} & \textbf{ATR} & \textbf{STU} & \textbf{FPD} & \textbf{OJR} & \textbf{Avg.} & \textbf{EPM} & \textbf{ASI} & \textbf{HLD} & \textbf{Avg.} & \textbf{REC} & \textbf{SSR} & \textbf{CRR} & \textbf{Avg.} & \\
\midrule
\rowcolor[gray]{0.92}
\multicolumn{20}{c}{\textbf{Traditional Streaming Prefill (Baseline)}} \\
Qwen2.5-VL-3B & $\infty$ & $\infty$ & -- & 67.11 & 44.04 & 64.66 & 40.45 & 65.35 & 58.70 & 56.72 & 51.85 & 54.05 & 17.20 & 41.04 & 27.37 & 34.75 & 40.00 & 34.04 & 47.13 \\
Qwen2.5-VL-7B & $\infty$ & $\infty$ & -- & 67.11 & 47.71 & 70.43 & 43.82 & 71.29 & 60.33 & 60.11 & 56.23 & 60.14 & 10.22 & 42.19 & 26.33 & 38.16 & 40.43 & 34.97 & 49.35 \\
\midrule
\rowcolor[gray]{0.92}
\multicolumn{20}{c}{\textbf{ViCoStream\textsuperscript{\textdagger}}} \\
\multirow{8}{*}{Qwen2.5-VL-3B\textsuperscript{\textdagger}} & 4 & 64 & 0.25 & 63.76 & 45.87 & 62.93 & 40.45 & 64.36 & 57.07 & 55.74 & 47.81 & 51.35 & 12.90 & 37.36 & 23.59 & 33.47 & 40.00 & 32.35 & 45.30 \\
& 8 & 64 & 0.25 & 65.77 & 49.54 & 62.93 & 38.20 & 63.37 & 57.07 & 56.15 & 51.18 & 54.73 & 17.20 & 41.04 & 21.64 & 34.58 & 40.00 & 32.07 & 46.35 \\
& 4 & 64 & 0.1 & 52.35 & 40.37 & 55.17 & 40.45 & 64.36 & 50.00 & 50.45 & 48.15 & 51.35 & 4.84 & 34.78 & 25.30 & 33.92 & 40.00 & 33.07 & 42.19 \\
& 8 & 64 & 0.1 & 50.34 & 44.04 & 58.62 & 39.89 & 62.38 & 51.09 & 51.06 & 47.14 & 53.38 & 18.28 & 39.60 & 19.17 & 35.08 & 40.00 & 31.41 & 43.28 \\
& 4 & 32 & 0.25 & 61.74 & 45.87 & 62.93 & 39.33 & 64.36 & 57.61 & 55.31 & 48.15 & 51.35 & 12.90 & 37.47 & 24.42 & 33.21 & 40.00 & 32.54 & 45.16 \\
& 8 & 32 & 0.25 & 65.10 & 46.79 & 63.79 & 39.33 & 65.35 & 57.61 & 56.33 & 48.15 & 54.73 & 16.67 & 39.85 & 21.87 & 34.68 & 40.00 & 32.18 & 46.17 \\
& 4 & 32 & 0.1 & 51.68 & 41.28 & 57.76 & 39.89 & 65.35 & 50.54 & 51.08 & 47.81 & 52.03 & 5.38 & 35.07 & 23.42 & 34.64 & 40.00 & 32.69 & 42.48 \\
& 8 & 32 & 0.1 & 50.34 & 44.95 & 58.62 & 39.33 & 63.37 & 53.80 & 51.73 & 49.49 & 53.38 & 18.82 & 40.56 & 18.78 & 35.28 & 40.00 & 31.35 & 43.85 \\
\cmidrule{1-20}
\multirow{8}{*}{Qwen2.5-VL-7B\textsuperscript{\textdagger}} & 4 & 64 & 0.25 & 70.47 & 48.62 & 67.24 & 50.00 & 68.32 & 55.98 & 60.11 & 51.85 & 61.49 & 13.98 & 42.44 & 27.67 & 35.68 & 41.67 & 35.01 & 49.41 \\
& 8 & 64 & 0.25 & 76.51 & 48.62 & 62.93 & 48.31 & 69.31 & 60.87 & 61.09 & 56.57 & 62.84 & 19.89 & 46.43 & 28.36 & 35.07 & 40.00 & 34.48 & 50.77 \\
& 4 & 64 & 0.1 & 62.42 & 49.54 & 62.07 & 46.07 & 72.28 & 53.80 & 57.70 & 50.34 & 60.81 & 15.30 & 42.15 & 24.62 & 33.46 & 40.00 & 32.69 & 47.56 \\
& 8 & 64 & 0.1 & 65.10 & 47.71 & 62.93 & 44.38 & 70.30 & 52.17 & 57.10 & 53.20 & 59.46 & 11.83 & 41.50 & 24.90 & 34.82 & 40.00 & 33.24 & 47.23 \\
& 4 & 32 & 0.25 & 71.81 & 44.95 & 67.24 & 49.44 & 70.30 & 55.43 & 59.86 & 53.87 & 62.16 & 15.59 & 43.88 & 27.27 & 36.38 & 42.08 & 35.24 & 49.71 \\
& 8 & 32 & 0.25 & 75.84 & 46.79 & 64.66 & 46.07 & 69.31 & 58.70 & 60.23 & 55.56 & 62.16 & 21.51 & 46.41 & 28.19 & 35.40 & 40.00 & 34.53 & 50.35 \\
& 4 & 32 & 0.1 & 61.07 & 47.71 & 62.07 & 45.51 & 70.30 & 51.63 & 56.38 & 49.83 & 59.46 & 15.05 & 41.45 & 25.39 & 33.28 & 40.00 & 32.89 & 46.77 \\
& 8 & 32 & 0.1 & 63.09 & 45.87 & 62.83 & 43.60 & 74.26 & 53.25 & 57.15 & 52.94 & 60.81 & 12.37 & 42.04 & 24.25 & 34.63 & 40.00 & 32.96 & 47.32 \\
\bottomrule
\end{tabular}%
}
\caption{Expanded OvO-Bench results following the standard task order of Real-Time Visual Perception, Backward Tracing, and Forward Active Response.}
\label{tab:ovobench_detailed}
\vspace{-0.4cm}
\end{table*}

\begin{table*}[t]
\centering
\fontsize{6.8pt}{7.4pt}\selectfont
\setlength{\tabcolsep}{2.5pt}
\renewcommand{\arraystretch}{1.15}
\resizebox{\textwidth}{!}{%
\setlength{\aboverulesep}{0pt}
\setlength{\belowrulesep}{0pt}
\begin{tabular}{l c c c c c c c c c c c c}
\toprule
\textbf{Model} & $\mathit{acn}$ & $\bm{\textit{ur}}$ & $\mathit{s}$ & \textbf{ECA} $\uparrow$ & \textbf{EPM} (mIoU$\uparrow$ / F1$\uparrow$) & \textbf{GVQ} $\uparrow$ & \textbf{RAR} $\uparrow$ & \textbf{RVQ} $\uparrow$ & \textbf{TAL-PT} $\uparrow$ & \textbf{TAL-TH} $\uparrow$ & \textbf{TEM} $\uparrow$ & \textbf{TVG} $\uparrow$ \\
\midrule
\rowcolor[gray]{0.92}
\multicolumn{13}{c}{\textbf{Traditional Streaming Prefill (Baseline)}} \\
Qwen2.5-VL-3B & $\infty$ & $\infty$ & -- & 37.80 & 1.29 / 1.60 & 53.45 & 34.80 & 51.60 & 18.63 & 13.09 & 16.80 & 27.30 \\
Qwen2.5-VL-7B & $\infty$ & $\infty$ & -- & 45.00 & 1.66 / 2.10 & 56.55 & 36.60 & 54.60 & 20.17 & 12.61 & 26.70 & 33.35 \\
\midrule
\rowcolor[gray]{0.92}
\multicolumn{13}{c}{\textbf{ViCoStream\textsuperscript{\textdagger}}} \\
\multirow{4}{*}{Qwen2.5-VL-3B\textsuperscript{\textdagger}} & 4 & 64 & 0.25 & 37.20 & 1.58 / 2.05 & 52.41 & 35.80 & 48.20 & 21.74 & 14.57 & 20.80 & 32.30 \\
& 4 & 64 & 0.1 & 37.20 & 1.62 / 1.90 & 54.83 & 36.00 & 46.20 & 20.84 & 14.41 & 16.45 & 30.00 \\
& 4 & 32 & 0.25 & 37.20 & 1.54 / 1.75 & 52.07 & 35.80 & 48.20 & 21.74 & 14.66 & 20.40 & 32.10 \\
& 4 & 32 & 0.1 & 37.20 & 1.62 / 1.95 & 54.48 & 35.20 & 45.60 & 21.63 & 14.42 & 16.35 & 30.35 \\
\cmidrule{1-13}
\multirow{4}{*}{Qwen2.5-VL-7B\textsuperscript{\textdagger}} & 4 & 64 & 0.25 & 37.60 & 2.02 / 2.55 & 53.79 & 36.60 & 51.20 & 17.93 & 13.54 & 21.90 & 28.05 \\
& 4 & 64 & 0.1 & 37.00 & 1.21 / 1.30 & 53.79 & 33.40 & 46.80 & 12.37 & 13.70 & 21.60 & 26.15 \\
& 4 & 32 & 0.25 & 37.60 & 2.04 / 2.50 & 54.48 & 35.60 & 51.20 & 17.95 & 12.51 & 21.80 & 28.30 \\
& 4 & 32 & 0.1 & 37.00 & 1.25 / 1.50 & 53.10 & 33.60 & 46.60 & 12.22 & 12.66 & 21.40 & 26.40 \\
\bottomrule
\end{tabular}%
}
\caption{Expanded ETBench task-level results. ECA, GVQ, RAR, and RVQ are accuracy-based tasks; EPM is temporal grounding with mIoU/F1; TAL-PT, TAL-TH, and TVG are F1-based temporal localization tasks; TEM reports mean recall.}
\label{tab:etbench_full}
\vspace{-0.4cm}
\end{table*}

\begin{table*}[htbp]
\centering
\small
\setlength{\tabcolsep}{4pt}
\renewcommand{\arraystretch}{1.15}

\setlength{\aboverulesep}{0pt}
\setlength{\belowrulesep}{0pt}
\begin{tabular}{c c c c c c c c c c c c c c}
\toprule
$\mathit{acn}$ & $\mathit{s}$ & \textbf{OP} & \textbf{CR} & \textbf{CS} & \textbf{ATP} & \textbf{EU} & \textbf{TR} & \textbf{PR} & \textbf{SU} & \textbf{ACP} & \textbf{CT} & \textbf{All} \\
\midrule
\rowcolor[gray]{0.92}
\multicolumn{13}{c}{\textbf{Traditional Streaming Prefill}} \\
\multicolumn{2}{c}{ Baseline} & 74.53 & 75.78 & 74.13 & 76.92 & 74.84 & 74.45 & 70.37 & 63.41 & 66.19 & 46.81 & 70.32 \\
\midrule
\rowcolor[gray]{0.92}
\multicolumn{13}{c}{\textbf{Frame-Level ViCoStream (c=1)}} \\
16 & 0.25 & 73.98 & 70.31 & 74.76 & 76.60 & 76.73 & 70.09 & 64.81 & 64.63 & 65.06 & 40.43 & 68.80 \\
16 & 0.1  & 71.00 & 71.88 & 73.82 & 71.15 & 72.33 & 69.47 & 64.81 & 64.23 & 66.19 & 28.19 & 66.48 \\
\midrule
\rowcolor[gray]{0.92}
\multicolumn{13}{c}{\textbf{Chunk-Level ViCoStream (c=4)}} \\
4 & 0.25 & 73.17 & 69.53 & 75.39 & 77.88 & 70.44 & 74.45 & 72.22 & 63.41 & 69.60 & 51.06 & 70.68 \\
4 & 0.1  & 69.92 & 70.31 & 74.45 & 76.28 & 72.96 & 69.78 & 70.37 & 62.20 & 68.47 & 39.89 & 68.28 \\
8 & 0.25 & 73.44 & 71.88 & 76.66 & 77.24 & 74.84 & 74.45 & 69.44 & 63.01 & 67.05 & 48.40 & 70.48 \\
8 & 0.1  & 68.83 & 69.53 & 76.03 & 76.28 & 71.70 & 67.60 & 68.52 & 63.41 & 65.91 & 40.96 & 67.68 \\
\bottomrule
\end{tabular}
\caption{Expanded StreamingBench per-task accuracy of frame-level  dropping on Qwen2.5-VL-3B-Instruct, compared with the chunk-level ViCoStream configurations of the same model reported in \Cref{tab:frame_level_dropping_summary}. The frame-level setting uses chunk size $c{=}1$ with full-frame inputs and a per-frame temporal-similarity threshold, so each frame is treated as an independent token-dropping unit. The chunk-level settings reuse the chunked prefilling and bounded attention from \Cref{sec:coordinated_design}. All variants share the same backbone, training data, and evaluation protocol.}
\label{tab:frame_level_dropping}
\vspace{-0.3cm}
\end{table*}

\FloatBarrier


\end{document}